\documentclass[runningheads]{llncs}
\usepackage{graphicx}

\usepackage{tikz}
\usepackage{comment}
\usepackage{amsmath,amssymb} 
\usepackage{color}
\usepackage{gensymb}
\usepackage[caption=false]{subfig}
\usepackage{multirow}
\usepackage[american]{babel}
\usepackage{makecell}
\usepackage{overpic}
\usepackage{ragged2e}

\usepackage{bbm}
\usepackage[breaklinks=true,bookmarks=false]{hyperref}
\usepackage{tabularx}
\usepackage{booktabs}
\usepackage{relsize}
\usepackage{wrapfig}
\usepackage{floatrow}
\usepackage[accsupp]{axessibility}  

\newcommand{\myparagraph}[1]{\vspace{2pt}\noindent{\bf #1}}
\newfloatcommand{capbtabbox}{table}[][\FBwidth]

\newcolumntype{P}[1]{>{\centering\arraybackslash}p{#1}}
\newcolumntype{M}[1]{>{\centering\arraybackslash}m{#1}}
\newcolumntype{L}[1]{>{\raggedright\arraybackslash}m{#1}}
\usepackage{xspace}

\makeatletter
\DeclareRobustCommand\onedot{\futurelet\@let@token\@onedot}
\def\@onedot{\ifx\@let@token.\else.\null\fi\xspace}

\def\eg{\emph{e.g}\onedot} 
\def\ie{\emph{i.e}\onedot} 
\def\cf{\emph{c.f}\onedot}

\def\etal{\emph{et al}\onedot}

\usepackage{mathtools}

\begin{document}
\pagestyle{headings}
\mainmatter
\def\ECCVSubNumber{1118}  



\title{CycDA: Unsupervised Cycle Domain Adaptation\\to Learn from Image to Video}

\titlerunning{CycDA: Unsupervised Cycle Domain Adaptation from Image to Video}
%

\author{Wei Lin\inst{1,5} \and
Anna Kukleva\inst{2} \and
Kunyang Sun\inst{1,3} \and
Horst Possegger\inst{1} \and
Hilde Kuehne\inst{4} \and
Horst Bischof\inst{1} }
\authorrunning{W. Lin et al.}
%
\institute{Institute of Computer Graphics and Vision, Graz University of Technology, Austria \\
\email{\{wei.lin,possegger,bischof\}@icg.tugraz.at}\\ 
\and
Max-Planck-Institute for Informatics, Germany \ \email{akukleva@mpi-inf.mpg.de}\\
\and
Southeast University, China \  \email{sunky@seu.edu.cn}
\and
Goethe University Frankfurt, Germany \email{kuehne@uni-frankfurt.de}
\and
Christian Doppler Laboratory for Semantic 3D Computer Vision 
}

\maketitle

\begin{abstract}
Although action recognition has achieved impressive results over recent years, both collection and annotation of video training data are still time-consuming and cost intensive. 
Therefore, image-to-video adaptation has been proposed to exploit labeling-free web image source for adapting on unlabeled target videos. This poses two major challenges: (1) spatial domain shift between web images and video frames; (2) modality gap between image and video data. 
To address these challenges, we propose Cycle Domain Adaptation (CycDA), a cycle-based approach for unsupervised image-to-video domain adaptation. We leverage the joint spatial information in images and videos on the one hand and, on the other hand, train an independent spatio-temporal model to bridge the modality gap. 
We alternate between the spatial and spatio-temporal learning with knowledge transfer between the two in each cycle. 
We evaluate our approach on benchmark datasets for image-to-video as well as for mixed-source domain adaptation achieving state-of-the-art results and demonstrating the benefits of our cyclic adaptation. Code is available at \url{https://github.com/wlin-at/CycDA}.

\end{abstract}

\section{Introduction}

The task of action recognition has seen tremendous success in recent years with top-performing approaches typically requiring large-scale labeled video datasets \cite{feichtenhofer2020x3d,wang2021action,yang2020temporal}, which can be impractical in terms of both data collection and annotation effort. In the meanwhile, webly-supervised learning has been explored to leverage the large amount of easily accessible web data as a labeling-free data source for video recognition \cite{gan2017deck,guo2018curriculumnet,li2017attention,wang2017untrimmednets,yang2018recognition,zhuang2017attend}.

In this work, we address the problem of image-to-video adaptation with webly-labeled images as the source domain and unlabeled videos as the target domain to allow for action classification without video annotation. 
This setting provides two major challenges: (1) the spatial domain shift between web images and video frames, based on difference in image styles, camera perspectives and semantic drifts; (2) the modality gap between spatial images and spatio-temporal videos. 
Specifically, this modality gap restrains that merely spatial knowledge can be transferred from source to target domain. 
%
%
%
Previous works on action recognition with web supervision either learn from web data directly~\cite{gan2016webly,gan2016you} or perform joint training by combining the web source with annotated target data~\cite{duan2020omni,ma2017less}. 
To specifically address the domain shift between web images and target videos, some approaches perform class-agnostic domain-invariant feature learning either within~\cite{kae2020image} or across modalities~\cite{yu2018exploiting,yu2019exploiting,liu2020deep}, in the absence of ensuring domain-invariance on the category-level. 

\begin{figure}[!t]
\centering
\includegraphics[width=\textwidth]{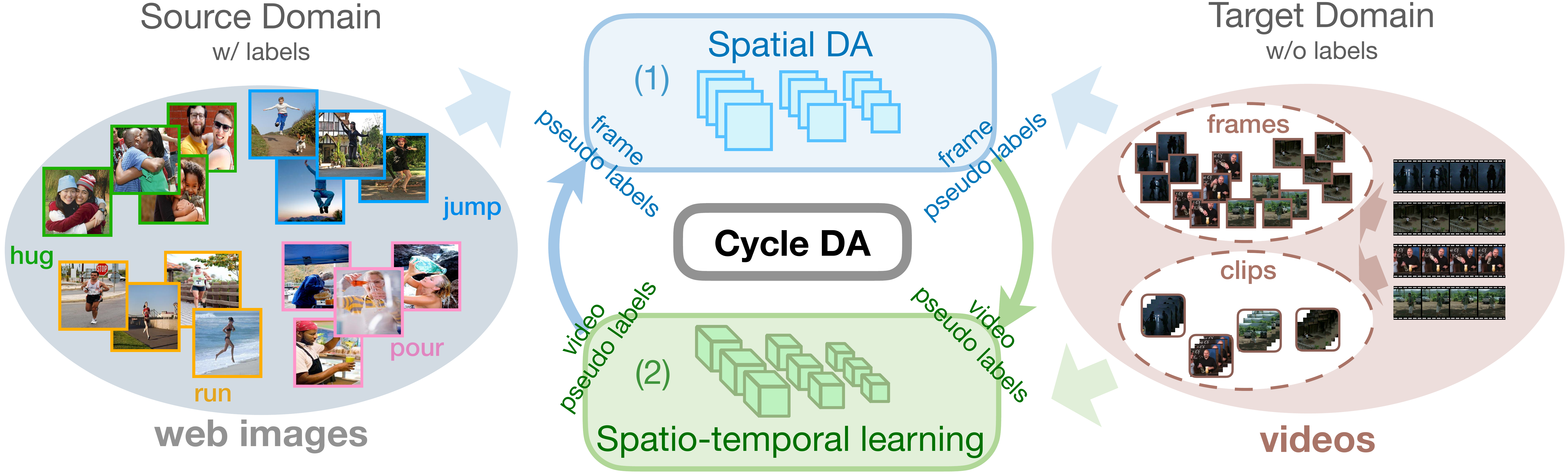}
\caption{Cycle Domain Adaptation (CycDA) pipeline: 
we address image-to-video adaption by training a spatial model and a spatio-temporal model alternately, passing pseudo labels to supervise each other in a cycle. The two alternating steps are: (1) domain alignment on the spatial model with pseudo labels from the spatio-temporal model, and (2) training the spatio-temporal model with updated pseudo labels from the spatial model.}
\label{fig:teaser_fig}
\end{figure}

In this context, we propose Cycle Domain Adaptation (CycDA), \ie alternating knowledge transfer between a spatial model and a spatio-temporal model. Compared to other works, we address the two challenges at hand, domain-alignment and closing the modality gap in separate stages, cycling between both of them. An overview of the CycDA is given in Fig.~\ref{fig:teaser_fig}. With the category knowledge from the spatio-temporal model, we achieve enhanced category-level domain invariance on the spatial model. With updated knowledge transferred from the spatial model, we attain better spatio-temporal learning. In this manner, we can better tackle each challenge for the corresponding model, with the updated knowledge transferred from the other. 

More specifically, we propose a four stage framework to address the domain shift between images and videos on different levels. 
In stage 1, we enforce \textit{class-agnostic} domain alignment on the spatial model between images and video frames.
In stage 2, we use supervision from the spatial model to learn a spatio-temporal video model, bridging the gap between the two modalities.
Stage 3 then focuses on \textit{class-aware} domain alignment on the spatial model, given pseudo labels computed by the video model trained on stage 2. 
In stage 4, we update the video model with the improved pseudo labels from the spatial model of stage 3.

We first evaluate our approach on several challenging settings for web image based action recognition, where a single cycle already outperforms baselines and state-of-the-arts.
Second, we show how CycDA can be flexibly applied for mixed-source image\&video-to-video DA settings, leading to a performance competitive to the state-of-the-art requiring only 5\% of the provided source videos.

We summarize our contributions as follows:
(1) We propose to address web image-to-video domain adaptation by decoupling the domain-alignment and spatio-temporal learning to bridge the modality gap.
(2) We propose cyclic alternation between spatial and spatio-temporal learning to improve spatial and spatio-temporal models respectively.
(3) We provide an extensive evaluation with different benchmark tasks that shows state-of-the-art results on unsupervised image-to-video domain adaptation and a competitive performance for the mixed-source image\&video-to-video setting.

\section{Related Work}

\textbf{Webly-supervised action recognition}. Various works have shown how web images and videos can be used as labeling-free data source to improve the performance of action recognition~\cite{duan2020omni,gan2016webly,gan2016you,ma2017less}. Gan \etal~\cite{gan2016webly,gan2016you} train with web data only, without adaptation to the target domain. 
Ma \etal~\cite{ma2017less} combine web images and video frames to train a spatial model and achieve comparable performance by replacing parts of annotated videos with web images. Duan \etal~\cite{duan2020omni} transform different types of web modalities to trimmed video clips and combine them with labeled target videos for joint training. In our work, we specifically address the domain shift on the spatial level and transfer the knowledge to the spatio-temporal level, without using any annotations on target data. 

\textbf{Image-to-video DA}. Compared to webly-supervised learning, image-to-video DA approaches actively address the domain shift between web images and target videos either by spatial alignment between web images and video frames \cite{li2017attention,sun2015temporal,zhang2016semi}, or through class-agnostic domain-invariant feature learning for images and videos~\cite{liu2020deep,yu2019exploiting,yu2018exploiting}. Li \etal~\cite{li2017attention} use a spatial attention map for cross-domain knowledge transfer from web images to videos. In this case, the DA is addressed on the spatial level without transfer to the temporal level. Liu \etal~\cite{liu2020deep} perform domain-invariant representation learning for images, video keyframes and videos, and fuse features of different modalities. Furthermore, hierarchical GAN~\cite{yu2018exploiting}, symmetric GAN~\cite{yu2019exploiting} and spatio-temporal causal graph~\cite{chen2021spatial} are proposed to learn the mapping between image features and video features. 
Closest to our work is probably the work of Kae \etal~\cite{kae2020image} which also employs a spatial and a spatio-temporal model for two stages of class-agnostic domain alignment, proposing to copy the weights from the spatial to the spatio-temporal model.

In contrast to these, we propose to transfer knowledge in the form of pseudo labels, without enforcing spatial information from the DA stage onto the spatio-temporal model.
Moreover, we alternately conduct spatial alignment and spatio-temporal learning with knowledge transfer to each other. With category knowledge transferred to the spatial model, we perform class-aware domain alignment that induces domain-invariance within category-level. 



\textbf{Video-to-video DA}. Compared to image-to-video adaption, video-to-video adaptation methods adapt annotated source videos to unlabeled target videos \cite{chen2019temporal,sahoo2021contrast,kim2021learning,choi2020shuffle,choi2020unsupervised,munro2020multi,luo2020adversarial,pan2020adversarial,jamal2018deep}, focusing mainly on the problem of domain alignment. Chen \etal \cite{chen2019temporal} align the features spatially on the frame-level and temporally on the scale-level. Others propose feature alignment via self-attention~\cite{choi2020shuffle}, cross-domain co-attention~\cite{pan2020adversarial}, or across two-stream modalities of RGB and optical flow~\cite{kim2021learning,munro2020multi}. Sahoo \etal~\cite{sahoo2021contrast} propose temporal contrastive learning and background mixing for domain-invariance. 
In this work, we focus on image-to-video adaptation and use only single stream of RGB.
We further show that we are able to extend the pipeline to the mixed-source case, where we achieve competitive performance compared to video-to-video adaptation methods while requiring only a small amount of source videos.

\section{Cycle Domain Adaptation (CycDA)}

\begin{figure}[!t]
\centering
\includegraphics[width=\textwidth]{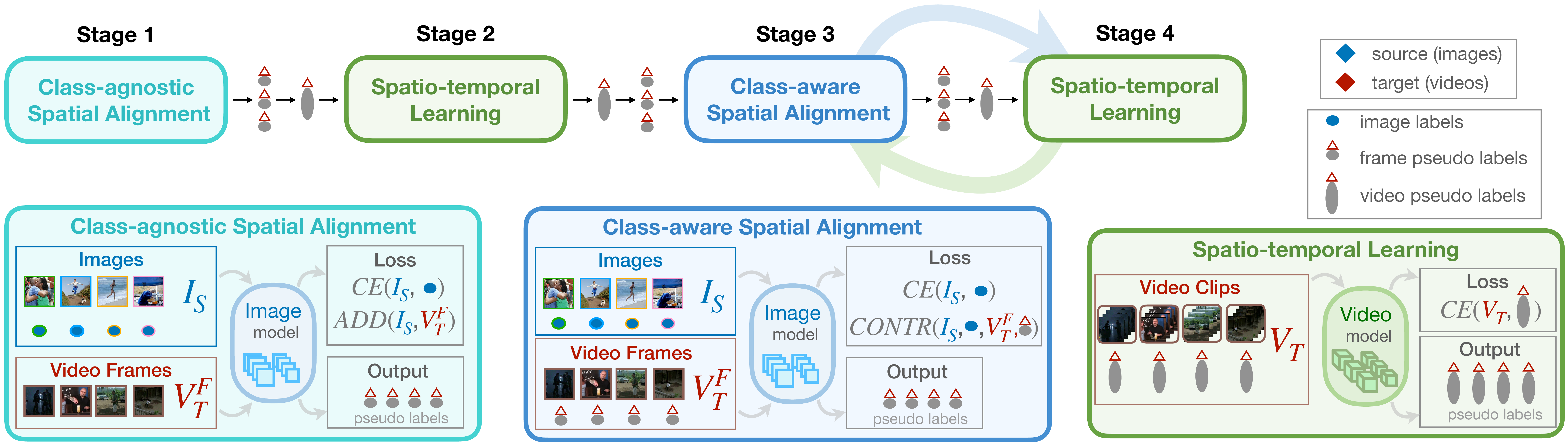}
\caption{Our CycDA framework alternates between spatial alignment (stage 1 and 3) and spatio-temporal learning (stage 2 and 4). See text for details.}
\label{fig:method_pipeline}
\end{figure}
We propose four stages to tackle image-to-video adaptation, which we summarize in Sec.~\ref{sec:overview}. 
Afterwards, we detail each stage and motivate the cycling of stages in Sec.~\ref{subsec:stages}. Our CycDA can be flexibly extended (Sec.~\ref{subsec:mixsource}) for mixed-source video adaptation, where a limited amount of annotated source videos are available.

\subsection{System Overview}
\label{sec:overview}
The task of unsupervised image-to-video DA is to learn a video classifier given labeled source images and unlabeled target videos. 
In order to close the domain gap across these two different modalities, we employ (1) a spatial (image) model to train on source web images and frames sampled from target videos, and (2) a spatio-temporal (video) model to train on target video clips.  
We propose a training pipeline that alternately adapts the two models by passing pseudo labels to supervise each other in a cycle. This facilitates the knowledge transfer between both models, where pseudo labels efficiently guide the model through the corresponding task, \ie semantic alignment (image model) or spatio-temporal learning (video model).

As shown in Fig.~\ref{fig:method_pipeline}, our CycDA pipeline consists of four training stages. 
The initial pseudo labels from \textit{class-agnostic} domain alignment (Stage 1) are improved via video spatio-temporal learning (Stage 2). In Stage 3, these pseudo labels on target data, together with ground truth labels on source, enable semantic alignment through the class-aware domain alignment. This again refines the pseudo labels, which further provide enhanced supervision for spatio-temporal learning in stage 4. In this manner, one iteration of CycDA facilitates the alternating knowledge transfer between the image and video models, whereby pseudo labels are improved on each stage and further provide strengthened supervision for better learning on the next stage. 



\myparagraph{Notations:} First, we denote the feature extractor as $E (\cdot; \theta_E)$, the classifier as $C(\cdot; \theta_C)$, and the domain discriminator as $D(\cdot;\theta_D)$. Then, we have the image model $\phi^I = \{E^I (\cdot; \theta^I_E), \allowbreak \ C^I (\cdot; \theta^I_C), \ D^I (\cdot; \theta^I_D)\}$ and the video model $\phi^V = \{E^V (\cdot; \theta^V_E), \ C^V (\cdot; \theta^V_C)\}$. 
We use the superscripts $I$, $V$ and $F$ to denote modalities of image, video and video frame, correspondingly. $S$ and $T$ stand for \textit{source} and \textit{target} domains respectively. The labeled source image domain is denoted as $I_S = \{( i_j, l(i_j)) |^{N^I_S}_{j=1} \}$, where $l(\cdot)$ is the ground truth label of the corresponding image. The unlabeled target video domain is $V_T = \{v_j |^{N^V_T}_{j=1} \}$ and each video $v_j$ has $M_j$ frames, the set of frames of unlabeled target videos $V^F_T = \{ \{ v^F_{j,m} |^{M_j}_{m=1} \} |^{N^V_T}_{j=1} \}$. 


\subsection{Stages}
\label{subsec:stages}
\textbf{Stage 1 - Class-agnostic Spatial Alignment}.
In the first stage, we learn the class-agnostic domain alignment between source web images and frames sampled from unlabeled target videos. 
Thus, we reduce the domain gap between the appearance of the web images and target videos even if the classes could be incorrectly aligned during this stage. 
We train the image model $\phi^I$ with a supervised \textbf{c}ross \textbf{e}ntropy loss $\mathcal{L}_{CE}( I_S )$ and an \textbf{a}dversarial \textbf{d}omain \textbf{d}iscrimination loss $\mathcal{L}_{ADD}( I_S, V^F_T )$ on source images and target frames.
With the classification loss on source images given as  
\begin{equation}
    \mathcal{L}_{CE}( I_S )=\sum_{(i_j, l(i_j))\in I_S} 
    -l(i_j) \cdot \log (C^I(E(i_j; \theta^I_E) ; \theta^I_C ))
\end{equation}
and the binary cross entropy loss for domain discrimination given as  
\begin{equation}
    \mathcal{L}_{ADD}(I_S, V^F_T )=\sum_{i_j, v^F_{j',m}} 
    \log D^I(E^I(i_j; \theta^I_E) ; \theta^I_D ) 
    + \log (1 - D^I(E^I(v^F_{j',m}; \theta^I_E) ; \theta^I_D ) ),
\end{equation}
the overall objective is 
$\min_{\theta^I_E, \theta^I_C} \mathcal{L}_{CE}( I_S ) + \beta \max_{\theta^I_E} \min_{\theta^I_D}  \mathcal{L}_{ADD}( I_S, V^F_T )$,
where $\beta$ is the trade-off weight between the two losses. We train the domain discriminator $D^I$ to distinguish between extracted features from different domains, while the feature extractor is trained adversely based on the domain discrimination task. In this case, the feature extractor learns to yield domain invariant features that the domain discriminator is unable to differentiate. The adversarial training is performed with a gradient reversal layer (GRL)~\cite{ganin2015unsupervised,ganin2016domain} which reverses the sign of gradients of the domain discrimination loss on the feature extractor during back-propagation. 
The domain alignment on this stage is class-agnostic as there is not yet any pseudo label for category knowledge in the target domain. The alignment is performed globally at the domain level. 

\myparagraph{Stage 2 \& Stage 4 - Spatio-Temporal Learning}.
In this stage, we use the trained image model $\phi^I$ from the previous stage to generate pseudo labels for the target videos. Then we perform supervised spatio-temporal learning with the pseudo labeled target data. 

Specifically, we first use $\phi^I$ to predict the pseudo label $\hat{l}(\cdot)$ for each frame of the target videos. We employ a spatio-temporal model that trains on target videos capturing both spatial and temporal information in the target domain only. To select new pseudo label candidates, we temporally aggregate frame-level predictions into a video-level prediction. We discard predictions with confidence lower than a threshold $\delta_p$ and perform a majority voting among the remaining predictions to define the video label. From all videos, we only keep those that have at least one frame with a minimum confidence. We set the confidence threshold $\delta_p$ such that $p\times100$\% of videos remain after the thresholding. 



We denote the target video set after thresholding as $\tilde{V}_T$. For stage 2, the supervised task on pseudo labeled target videos is to $\min_{\theta^V_E, \theta^V_C} \hat{\mathcal{L}}_{CE}( \tilde{V}_T )$, with
\begin{equation}
    \hat{\mathcal{L}}_{CE}( \tilde{V}_T )=\sum_{v_j\in \tilde{V}_T} 
    -\hat{l}(v_j) \cdot \log (C^V(E^V(V_j; \theta^V_E) ; \theta^V )).  
\end{equation}  
In stage 4, we repeat the process as described above and re-train the video model on the target data with the updated pseudo labels from the third stage.

\myparagraph{Stage 3 - Class-aware Spatial Alignment}.
The adversarial learning for domain discrimination in the first stage aligns features from different domains globally, but not within each category (\cf Fig.~\ref{fig:tsne_plots} (b) and (c)). In this case, target samples in a category A can be incorrectly aligned with source samples in a different category B. This would lead to inferior classification performance of the target classifier. To evade this misalignment, we perform class-aware domain alignment in the third stage between the source web images and the target video frames.
Since the source data consists exclusively of images, we apply alignment on the spatial model between images and frames.
Furthermore, as the target data is unlabeled, in order to align features across both domains within each category, we generate pseudo labels by the model $\phi^V$ from the second stage to provide category knowledge. Specifically, we use the video model to generate video-level labels that we disseminate into frame-level labels. 
To align images and video frames we use cross-domain contrastive learning by maximizing the similarity between samples across domains of the same class and minimizing the similarity between samples from different classes. 
We use $z = E^I(i; \theta^I_E)$ to denote the feature computed by the feature extractor on image $i$. The set of source image features is $Z^I_S = \{ E^I(i; \theta^I_E )| i \in I_S \}$ and the set of target frame features is $Z^F_T = \{ E^I(v^F; \theta^I_E )| v^F \in V^F_T \}$. 
During training, for each pseudo labeled target sample $z^F_j\in Z^F_T$, we randomly choose two samples from the source domain: a positive sample of the same label and a negative sample of a different label, \ie $z^I_{j \ +}, z^I_{j \ -} \in I_S$. The contrastive loss is formulated as
\begin{equation} 
    \mathcal{L}_{CONTR}(I_S, V^F_T )=
     -\sum_{  z^F_j\in Z^F_T  }  
     \log \frac{ h( z^F_j, z^I_{j \ +} ) }{ h( z^F_j, z^I_{j \ +} ) +  h( z^F_j, z^I_{j \ -} ) }  .
\end{equation} 
Following \cite{chen2020simple},  we set $h(u, v) = \exp( \text{sim}(u, v) / \tau)$, where we use the cosine similarity $\text{sim}(u, v)=u^\mathsf{T}v / (\lVert u \rVert \lVert v \rVert)$ and $\tau$ is the temperature parameter. Thus, the objective of stage 3 on the image model is $\min_{\theta^I_E, \theta^I_C} \mathcal{L}_{CE}( I_S ) +  \mathcal{L}_{CONTR}( I_S, V^F_T )$.

In the third stage, an alternative of exploiting pseudo labels from the video model from the second stage is to self-train the video model on the target data, as self-training is a common practice in DA~\cite{liu2021cycle,zhang2020label,zou2018unsupervised,zou2019confidence}.  
However, the category-level domain alignment with supervision from the source domain further regularizes the learning of class distribution in the target domain. This results in a significantly improved target classifier, as we show in Sec.~\ref{sec:ablation_study}.

\myparagraph{Cycling of the Stages}.
The pseudo labels from the video model are exploited for class-aware domain alignment on the image model (stage 3) and the updated pseudo labels from the image model can supervise the training of the video model (stage 4). In this manner, stage 3 and stage 4 can be performed iteratively. We show in the evaluation (Table~\ref{tab:results_e2h_s2h_b2u} and Fig.~\ref{fig:n_iters_vs_performance_e2h}) the impact of this cyclic learning setup and how several iterations of CycDA can further improve the performance of the target classifier. 


\subsection{Mixed-source Video Adaptation} 
\label{subsec:mixsource}
Image-to-video DA applies to the case in which the source domain consists only of web images. However, other possible settings presume limited amount of annotated videos with the domain shift to the unlabeled target videos. We refer to this case as mixed-source video adaptation.
CycDA can be adjusted for this setting as follows.
We denote the labeled source video domain as $V_S=\{( v_j, l(v_j)) |^{N^V_S}_{j=1} \}$. 
For the class-agnostic (stage 1) and class-aware domain alignment (stage 3) stages we replace the source image domain $\{I_S\}$ by the mixed-source domain data $\{I_S, F_S\}$ which consists of web images and frames sampled from source videos. The supervised classification, adversarial domain discrimination and cross-domain contrastive learning are adapted accordingly. For the spatio-temporal learning of the video model $\phi^V$ (stage 2 and 4) we include additional supervised classification w.r.t. the ground truth labels for the source videos, therefore the overall loss is $\mathcal{L}_{CE}( V_S ) + \hat{\mathcal{L}}_{CE}( \tilde{V}_T )$. In this case, the annotated source videos are utilized to regularize domain alignment on the image model, and provide further supervision for learning the classification task on the video model. In Sec.~\ref{sec:exp_mixed_source}, we demonstrate that in the context of mixed-source video adaptation, even a limited amount of source videos is sufficient to achieve results competitive to video-to-video adaptation approaches that employ the entire source video dataset.




\section{Experiments}
\subsection{Datasets}
To evaluate our CycDA framework for image-to-video adaptation, we conduct experiments on 3 real-world image-video action recognition benchmark settings. Videos are from two large-scale action recognition datasets, UCF101~\cite{soomro2012ucf101} and HMDB51~\cite{kuehne2011hmdb}. Web images are from the EADs (Extensive Action Dataset)~\cite{chen2021spatial}, Stanford40~\cite{yao2011human} and the BU101 dataset~\cite{ma2017less}. 

The three image-to-video adaptation tasks are: (1)~Stanford40 $\rightarrow$ UCF101: the UCF101 action dataset contains 13320 videos collected from YouTube with 101 action classes. The Stanford40 dataset contains 9532 images collected from Google, Bing and Flickr, comprised of 40 action classes. Following \cite{chen2021spatial,yu2019exploiting,yu2018exploiting}, we select the 12 common action classes between the two datasets for image-to-video action recognition. (2)~EADs$\rightarrow$HMDB51: HMDB51 has 6766 videos with 51 action classes collected from online videos and movie clips. The EADs dataset consists of Stanford40 and the HII dataset~\cite{tanisik2016facial}. It has 11504 images from 50 action classes. There are 13 shared action classes between the two datasets. (3)~BU101$\rightarrow$UCF101: BU101 consists of 23.8K web action images from 101 classes that completely correspond to classes on UCF101. We use data of all classes for evaluation on large-scale image-to-video adaptation. 

UCF101 and HMDB51 both have three splits of training and test sets. Following \cite{li2017attention,yu2018exploiting,yu2019exploiting}, we report the average performance over all three splits.

The UCF-HMDB dataset~\cite{chen2019temporal} is a benchmark for video-to-video DA. It consists of the 12 common classes between UCF101 and HMDB51. On this dataset, we perform two types of evaluations: (1) \textit{frame-to-video adaptation}: we use only a single frame from each source video to adapt to target videos; and (2) mixed-source video adaptation: we use source and target videos of UCF-HMDB, and extend the source domain with web images from BU101. 



\subsection{Implementation Details}
For the image model, we use a ResNet-18~\cite{he2016deep} pretrained on ImageNet~\cite{deng2009imagenet}. We freeze the first 6 sequential blocks and train with a learning rate of 0.001 to perform the domain alignment between web images and frames sampled from target videos. 
To avoid redundancy in the video frames, we uniformly divide a video into 5 segments. In each training epoch, we randomly sample one frame from each segment. As trade-off weight for domain discrimination, we follow the common practices in \cite{ganin2015unsupervised,ganin2016domain,chen2020action} to gradually increase $\beta$ from 0 to 1. The temperature parameter $\tau$ is set to 0.05.

For the video model, we employ I3D Inception v1~\cite{carreira2017quo} pretrained on the Kinetics dataset~\cite{kay2017kinetics}, which is common practice, \eg~\cite{chen2021spatial,choi2020shuffle,kim2021learning,munro2020multi,sahoo2021contrast}. We train the RGB stream only. To validate the efficacy of our CycDA pipeline, we use the I3D backbone with a shallow classifier of 2 FC layers, without any temporal aggregation module (\eg GCN in \cite{sahoo2021contrast} or self-attention module in \cite{choi2020shuffle}). We extract a clip of 64 frames from each video. Following \cite{sahoo2021contrast}, we use a learning rate of 0.001 for the backbone and 0.01 on other components. We keep $p=70\%$ and $80\%$ of videos in stage 2 and stage 4.

\subsection{Image-to-video DA}


\begin{table}[!tb]
\scriptsize
\centering
\begin{tabular}{M{4cm}M{3cm}M{1.5cm}M{1.5cm}M{1.5cm}} 
\toprule
Method & Backbone  &  E$\rightarrow$H &  S$\rightarrow$U & B$\rightarrow$U  \\
\midrule
source only & ResNet18  & 37.2 & 76.8 & 54.8 \\
\midrule
DANN~\cite{ganin2016domain}*  & ResNet18  & 39.6 & 80.3 & 55.3\\
UnAtt~\cite{li2017attention} & ResNet101 &  - & - &  66.4 \\
HiGAN~\cite{yu2018exploiting} & ResNet50, C3D & 44.6 &  95.4 & - \\ 
SymGAN~\cite{yu2019exploiting} & ResNet50, C3D  & 55.0  &   97.7 & - \\

CycDA (1 iteration) & ResNet50, C3D & 56.6 & 98.0 & - \\ 
\midrule
DANN~\cite{ganin2016domain}+I3D* & ResNet18, I3D & 53.8 & 97.9 & 68.3 \\
HPDA~\cite{chen2021spatial}* & ResNet50, I3D  & 38.2 & 40.0 & - \\
CycDA (1 iteration) & ResNet18, I3D  &  60.5 & 99.2 & 69.8 \\
CycDA (2 iterations) & ResNet18, I3D  &  60.3 & \textbf{99.3} & 72.1 \\
CycDA (3 iterations) & ResNet18, I3D  &  \textbf{62.0} & 99.1 & \textbf{72.6} \\
\midrule
supervised target & ResNet18, I3D & 83.2 & 99.3  & 93.1  \\
\bottomrule
\end{tabular}
\caption{
Results on E$\rightarrow$H (13 classes), S$\rightarrow$U (12 classes) and B$\rightarrow$U (101 classes), averaged over 3 splits. ResNet, C3D and I3D are pretrained on ImageNet\cite{he2016deep}, Sports-1M\cite{karpathy2014large} and Kinetics400\cite{kay2017kinetics}. * denotes our evaluation.}
\label{tab:results_e2h_s2h_b2u}
\end{table}



We compare the proposed approach to other image-to-video adaptation methods on the three described benchmark settings as shown in Table~\ref{tab:results_e2h_s2h_b2u}.
%
As CycDA enables the iterative knowledge transfer between the image model and the video model, we can repeat stage 3 and stage 4 multiple times.
We therefore report the performance for the first three iterations. 
We add the lower bound (source only) and the upper bound (ground truth supervised target) for reference.

We compare against several approaches: DANN~\cite{ganin2016domain} is classical adversarial domain discrimination on the image-level. UnAtt~\cite{li2017attention} applies a spatial attention map on video frames. HiGAN~\cite{yu2018exploiting} and SymGAN~\cite{yu2019exploiting} employ GANs for feature alignment (on backbone of ResNet50 and C3D) and define the current state-of-the-art on E$\rightarrow$H and S$\rightarrow$U. We also evaluate CycDA with the same backbones for fair comparison. DANN~\cite{ganin2016domain}+I3D is a strong baseline that trains the I3D model with pseudo labels from an adapted image model. HPDA~\cite{chen2021spatial} is a recent partial DA approach and for a fair comparison, we re-run its official implementation in a closed-set DA setting. 
Our CycDA outperforms all other approaches already after the first iteration. Except for the saturation on S$\rightarrow$U, running CycDA for more iterations leads to a further performance boost on all evaluation settings.

We further explore the potential of CycDA on UCF-HMDB, which is a benchmark for video-to-video adaptation. 
For a strict comparison, we select data from the same source video dataset used in the video-to-video adaptation methods, without using any auxiliary web data for training. However, instead of directly using the source videos, we perform \textit{frame-to-video} adaptation where we use only one frame from each source video to adapt to target videos. Here we sample the middle frame from each video and report the results in Table~\ref{tab:results_mixed_source_vid_adapt} (case B). We see that even when using only one frame per source video, on U$\rightarrow$H, CycDA (83.3\%) can already outperform TA\textsuperscript{3}N~\cite{chen2019temporal} (81.4\%) and SAVA~\cite{choi2020shuffle} (82.2\%) which use all source videos for video-to-video adaptation. This demonstrates the strength of CycDA to exploit the large informativity in single images such that they could potentially replace videos as the source data. On H$\rightarrow$U, our source domain contains only 840 frames from the 840 videos in the HMDB training set on UCF-HMDB, which leads to an inferior performance. 
We show that this can be easily addressed by adding auxiliary web data in Sec.~\ref{sec:exp_mixed_source}.




\begin{table}[!tb]
\scriptsize
\centering
\begin{tabular}{M{0.2cm}M{2.3cm}M{2.4cm}M{1.5cm}M{1cm}M{1.7cm}M{1cm}M{1cm}} 
\toprule
\multicolumn{2}{c}{\multirow{2}*{ DA setting}} & \multirow{2}*{Method} & \multirow{2}*{ \shortstack{Video\\backbone}}  & \multicolumn{2}{c}{Source data} & \multirow{2}*{U$\rightarrow$H} & \multirow{2}*{H$\rightarrow$U}  \\
\cmidrule{5-6}
~ & ~ & ~ & ~ & web image & videos (U or H) in \% & ~ & ~ \\
\midrule
\multirow{11}*{A:} & \multirow{11}*{video-to-video} & AdaBN~\cite{li2018adaptive} &  ResNet101  & - & 100\% & 75.5  &  77.4 \\
~ & ~ & MCD~\cite{saito2018maximum} &  ResNet101  & - & 100\% & 74.4  &  79.3 \\
~ &  ~ & TA\textsuperscript{3}N~\cite{chen2019temporal} & ResNet101  & - & 100\% & 78.3  &  81.8 \\
~ & ~ & ABG~\cite{luo2020adversarial} &  ResNet101  & - & 100\% & 79.1  &  85.1 \\
~ & ~ & TCoN~\cite{pan2020adversarial} &  ResNet101  & - & 100\% & 87.2  &  89.1 \\
~ & ~ & DANN~\cite{ganin2016domain} &  I3D & - & 100\%  & 80.7 & 88.0 \\
~ & ~ & TA\textsuperscript{3}N~\cite{chen2019temporal}  & I3D & - & 100\%  & 81.4 & 90.5 \\
~ & ~ & SAVA~\cite{choi2020shuffle}  &  I3D & - & 100\%  & 82.2 & 91.2 \\
~ & ~ & MM-SADA~\cite{munro2020multi} &  I3D & - & 100\% & 84.2  &  91.1 \\
~ & ~ & CrossModal~\cite{kim2021learning} & I3D & - & 100\% & 84.7  &  92.8 \\
~ & ~ & CoMix~\cite{sahoo2021contrast} &  I3D & - & 100\%  &86.7 & 93.9   \\
\midrule
B: & frame-to-video & CycDA &  I3D & - & one frame & 83.3   &  80.4 \\
\midrule
\multirow{5}*{C:} & \multirow{5}*{\shortstack{mixed-source\\to video}} & \multirow{5}*{CycDA} &  \multirow{5}*{I3D}   & BU* & 0\% & 77.8 & 88.6 \\
~ & ~ & ~& ~ & BU* & 5\% & 82.2 & 93.1 \\
~ & ~ & ~& ~ & BU* & 10\% & 82.5  & 93.5 \\
~ & ~ & ~& ~ & BU* & 50\% & 84.2 & 95.2 \\
~ & ~ & ~& ~ & BU* & 100\% & 88.1 & 98.0 \\
\midrule
\multicolumn{3}{c}{supervised target} & I3D & - & - & 94.4 & 97.0 \\
\bottomrule
\end{tabular}
\caption{Results of Cycle Adaption on UCF-HMDB in comparison to video-to-video adaptation (case A) approaches. For frame-to-video adaptation (case B), we use only one frame from each source video to adapt to target videos. For mixed-source video adaptation (case C), we combine BU101 web images and source videos as the source data. *We sample 50 web images per class from 12 classes in BU101. }
\label{tab:results_mixed_source_vid_adapt}
\end{table}

\subsection{Mixed-source image\&video-to-video DA}
\label{sec:exp_mixed_source}
For mixed-source adaptation, we assume to have both web images and some amount of source videos in the source domain. To evaluate this case, we use the source and target videos on UCF-HMDB, and extend the source domain with web images of the 12 corresponding action classes in BU101. We notice that using all web data from BU101 leads to performance saturation on the target video set of UCF-HMDB. Therefore, to validate the efficacy of CycDA, we only sample 50 web images per class as auxiliary training data. We vary the amount of source videos in the mixed-source domain and report the results in Table~\ref{tab:results_mixed_source_vid_adapt} (case C). First, by training with only sampled web images (without any source videos), we achieve baseline results of 77.8\% (BU$\rightarrow$H) and 88.6\% (BU$\rightarrow$U). By adding only 5\% of videos to the mixed-source domain, we already achieve performance comparable to the video-to-video adaptation methods, \ie 82.0\% (BU+U$\rightarrow$H) and 93.1\% (BU+H$\rightarrow$U). Furthermore, increasing the amount of source videos from 5\% to 50\% leads to another improvement of $2\%$. As web images are more informative than sampled video frames, using web images as auxiliary training data can thus significantly reduce the amount of videos required. 
Finally, with sampled web images and all source videos, we outperform all video-to-video adaptation methods, even exceeding the supervised target model  for BU+H$\rightarrow$U by $1\%$. Considering that we only use 50 web images per class, this further demonstrates that CycDA can exploit both, the information in web images and knowledge from the source data with domain shift, for a potentially improved learning.

\subsection{Ablation study}\label{sec:ablation_study}
We perform several ablation studies to validate the proposed CycDA pipeline. We conduct these experiments on the setting of EADs $\rightarrow$ HMDB51 (split 1). 

\begin{table}[!tb]
\scriptsize
\centering
\begin{tabular}{M{0.5cm}M{2.3cm}M{2.4cm}M{1.5cm}M{2.4cm}M{1.5cm}M{1cm}} 
\toprule
& Experiment &  stage 1 &  stage 2 &  stage 3 &  stage 4 &  Acc \\
\midrule
A: & source only & $\mathcal{L}_{CE}( I_S)$ &  - & - & - & 39.0 \\
\midrule
B: & source only + video model  & $\mathcal{L}_{CE}( I_S)$ &  $\hat{\mathcal{L}}_{CE}( V_T )$ & - & - &  50.5 \\
\midrule
C: & class-agnostic DA + video model &  $\mathcal{L}_{CE}( I_S)$, $\mathcal{L}_{ADD}( I_S, V^F_T )$ &  $\hat{\mathcal{L}}_{CE}( V_T )$ & - & -  & 52.3 \\  
\midrule
D: & case C + vid. self-train$\times$1 & $\mathcal{L}_{CE}( I_S)$, $\mathcal{L}_{ADD}(I_S, V^F_T )$ & $\hat{\mathcal{L}}_{CE}( V_T )$  &   $\hat{\mathcal{L}}_{CE}(V_T )$ &  - &  55.4 \\
\midrule
E: & case C + vid. self-train$\times$2 & $\mathcal{L}_{CE}( I_S)$, $\mathcal{L}_{ADD}( I_S, V^F_T )$ & $\hat{\mathcal{L}}_{CE}( V_T )$  &   $\hat{\mathcal{L}}_{CE}( V_T )$ &  $\hat{\mathcal{L}}_{CE}( V_T )$ & 56.4 \\
\midrule
F: & CycDA  & $\mathcal{L}_{CE}( I_S)$, $\mathcal{L}_{ADD}( I_S, V^F_T )$ & $\hat{\mathcal{L}}_{CE}( V_T )$ & $\mathcal{L}_{CE}( I_S)$, $\mathcal{L}_{CONTR}( I_S, V^F_T ) $  &  $\hat{\mathcal{L}}_{CE}( V_T )$ & \textbf{60.8} \\ 
\bottomrule
\end{tabular}
\caption{Stage-wise ablation study of the CycDA training pipeline on EADs $\rightarrow$ HMDB51 split 1. A: source only on image model. B: source only training on image model and video model training. C: class-agnostic DA (stage 1) and video model training (stage 2). D: case C + one stage of self-training the videos model. E: case C + two stages of self-training the video model. F: CycDA with stage 1$\sim$4. Category-level pseudo labels of case C and F are compared in Fig.~\ref{fig:ps_label_analysis}. 
}
\label{tab:stage_wise_ablation}
\end{table}

\textbf{Stage-wise ablation study}. We first validate the efficacy of the CycDA pipeline by switching the stages with alternate counterparts. We report the quantitative results of six ablation settings in Table~\ref{tab:stage_wise_ablation}. Case A (source only), training the image model on web images only and predicting on target test set, demonstrates the lower bound of 39\%. Case B (source only + video model), training with pseudo labeled target videos, is a vanilla baseline. It shows that training the video model with supervision from the image model already significantly improves performance by 11.5\%. In case C, with class-agnostic domain alignment on the image model, the performance of B is improved by 1.8\%. In case F, after completing the cycle with class-aware domain alignment in stage 3 and training the video model in stage 4 with the updated pseudo labels, we achieve the best performance with 60.8\%. 

Additionally, we conduct ablation experiments using pseudo labels from case C to self-train the video model. Although self-training the video model for 1 (case D) or 2 (case E) stages exhibits performance improvement compared to case C, it is still clearly outperformed by CycDA. This indicates that the elaborate step of knowledge transfer from the video model to the image model and class-aware domain alignment are critical for a good performance. 
\begin{figure}[!tb]
\centering
\includegraphics[width=0.95\textwidth]{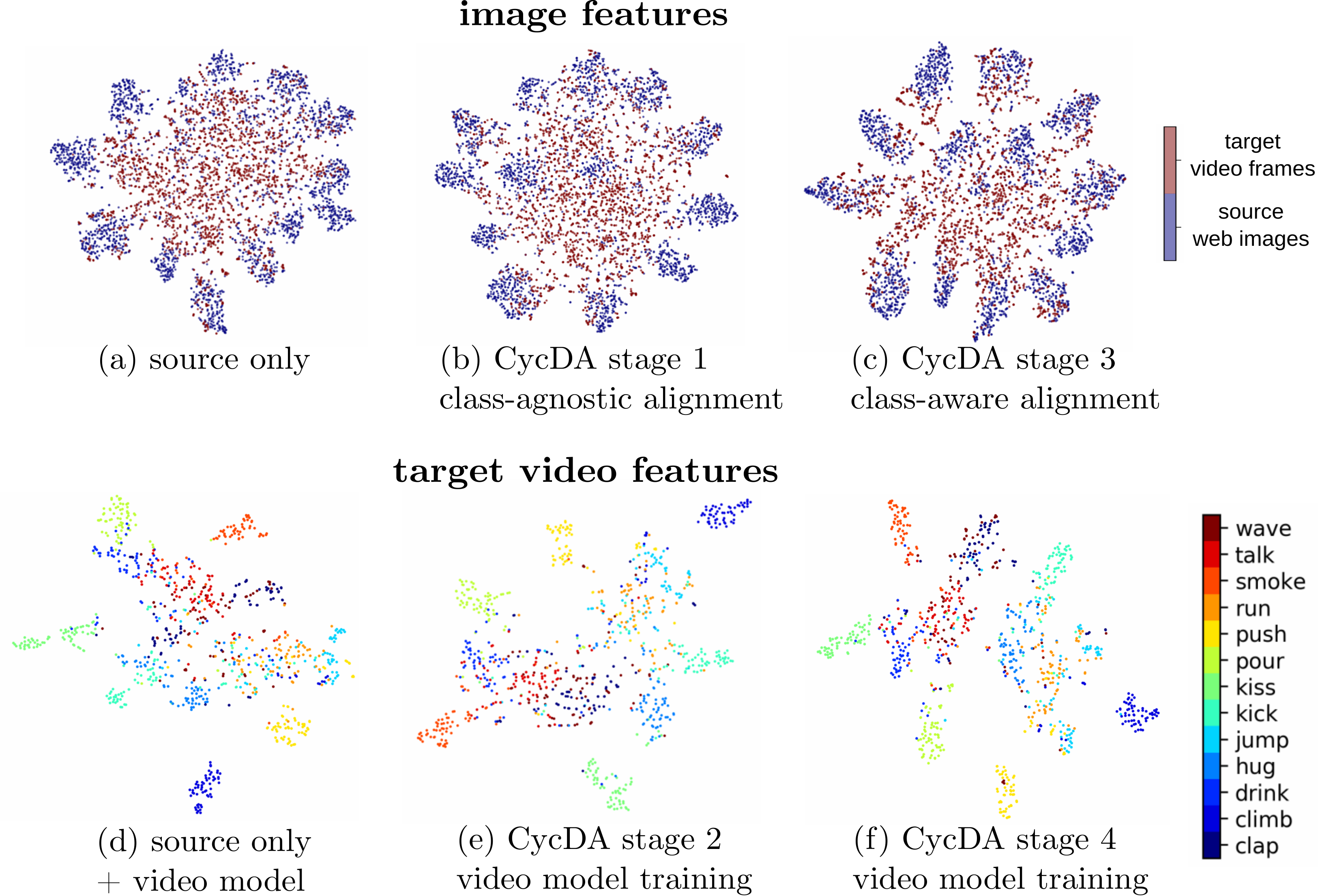}
\caption{t-SNE visualizations of image features (for both source and target) and target video features (colored w.r.t. ground truth). We plot the results of source only, source only and video model training, and results of each stage in CycDA. }
\label{fig:tsne_plots}
\end{figure}

We further illustrate the t-SNE~\cite{van2008visualizing} feature visualizations of ablation cases in Fig.~\ref{fig:tsne_plots}. By observing image features of the source only case (Fig.~\ref{fig:tsne_plots}(a)), we see that web images (blue) gather in category clusters after supervised training, while target video frames (red) are far less discriminative and highly misaligned with source due to large domain shift.
The class-agnostic domain alignment (Fig.~\ref{fig:tsne_plots}(b)) in stage 1 results in a slightly better global alignment of source and target features. With category knowledge from the video model, the class-aware domain alignment (Fig.~\ref{fig:tsne_plots}(c)) demonstrates distinctly better category-level association between source and target. By passing pseudo labels from the image model of Fig.~\ref{fig:tsne_plots}(a)(b)(c) to supervise the spatio-temporal training, we get the corresponding video models whose features are plotted in Fig.~\ref{fig:tsne_plots}(d)(e)(f). The vanilla baseline of source only + video model (Fig.~\ref{fig:tsne_plots}(d)) leads to highly undiscriminative features on the difficult classes, which is slightly improved by class-agnostic alignment (Fig.~\ref{fig:tsne_plots}(e)). Class-aware domain alignment (Fig.~\ref{fig:tsne_plots}(f)) results in more pronounced category clusters with larger inter-class distances.


\begin{figure}[!tb] 
\centering
\includegraphics[width=0.9\textwidth]{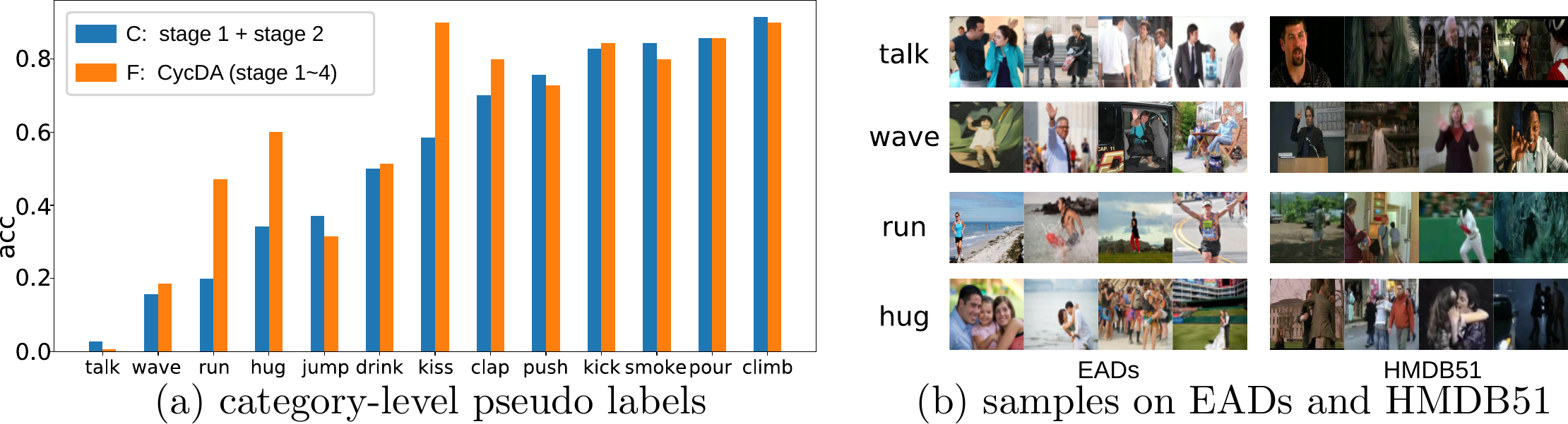}
\caption{(a) Category-level pseudo label analysis on target. We compare the accuracy of category-level pseudo labels on target videos for the ablation case C and F from Table~\ref{tab:stage_wise_ablation}. Here we plot the accuracy of pseudo labels of the 13 common classes on EADs $\rightarrow$ HMDB51. (b) Samples of four categories on EADs and HMDB51.
}
\label{fig:ps_label_analysis}
\end{figure}

\textbf{Category-level pseudo label analysis on target}. In Fig.~\ref{fig:ps_label_analysis}(a), we plot the accuracy of category-wise pseudo labels on target videos for the ablation cases C and F from Table~\ref{tab:stage_wise_ablation}. On case C (blue) which consists of stage 1 (class-agnostic spatial domain alignment) and stage 2 (spatio-temporal learning), image-to-video action recognition has varying performance on different action classes. More specifically, it yields better results on appearance-based actions with distinct background (\eg \textit{climb}), or on actions with discriminative pose or gesture (\eg \textit{smoke}, \textit{pour}). On the contrary, there is inferior performance on fine-grained actions with subtle movements (\eg \textit{talk}) or actions that are semantically highly generalized with large inter-class variation (\eg \textit{run}, \textit{wave}), \cf in Fig.~\ref{fig:ps_label_analysis}(b). 

By comparing the pseudo label accuracy of cases C and F, we see that the complete CycDA with stage 3 and stage 4 contributes to a significant performance boost on the difficult classes (\eg \textit{run, hug, kiss}) while keeping the performance on the easy classes. This can be attributed to the class-aware domain alignment that improves cross-domain association on the difficult classes while keeping the alignment on easy ones. Note that better class-aware domain alignment also leads to more support samples for difficult classes, which could result in slight performance drop on the easier classes.

\textbf{Domain adaptation strategies in stage 3}. In stage 3, we transfer knowledge from the video model to the image model by passing pseudo labels from the video model to provide category information in domain alignment. We compare different DA strategies that use pseudo labels from the video model in Table~\ref{tab:stage3_comparison}. Intuitively, performing the supervised task on both source and target (case B) outperforms the case of pseudo labeled target only (case A). Adding the adversarial domain discrimination (case C) leads to a further boost. The cross-domain contrastive learning (case D) has the best performance among the 4 cases. 
In comparison to the case of only stage 1 and 2, a complete cycle with 4 stages leads to performance improvement of at least 3.3\% for all cases. This indicates the benefits of transferring the knowledge from the video model onto the image model. The proposed CycDA generalizes well on different strategies using pseudo labels from the previous stage. 
\begin{table}[!tb]
\scriptsize
\centering
\begin{tabular}{M{2.8cm}|M{0.5cm}M{4cm}M{1.5cm}M{2.4cm}M{1.5cm}M{1cm}} 
\toprule
Experiment &   & stage 3 &  stage 4 &  Acc \\
\midrule
stage 1$\sim$2 & &  - & -  & 52.3 \\ 
\midrule
\multirow{4}*{stage 1$\sim$4} 
& A: & ${\mathcal{L}}_{CE}( V^F_T )$ 
&  \multirow{4}*{ $\hat{\mathcal{L}}_{CE}( V_T )$} 
& 55.6 \\ 
~ &  B:  &$\mathcal{L}_{CE}( I_S, V^F_T  )$ &  ~ &  56.4 \\
~ &   C:  &$\mathcal{L}_{CE}( I_S, V^F_T  )$, $\mathcal{L}_{ADD}( I_S, V^F_T )$  &  ~ & 58.0 \\
~ &    D:  & $\mathcal{L}_{CE}( I_S)$, $\mathcal{L}_{CONTR}( I_S, V^F_T ) $ &  ~ & \textbf{60.8} \\

\bottomrule
\end{tabular}
\caption{Comparison of DA strategies in stage 3 on EADs $\rightarrow$ HMDB51 split 1. A: supervised classification on pseudo labeled target frames. B: supervised classification on source and pseudo labeled target. C: B + adversarial domain discrimination. D: supervised classification on source + cross-domain contrastive learning. 
}
\label{tab:stage3_comparison}
\end{table}

\begin{figure}[!tb]
\begin{floatrow}
\capbtabbox{%
  \scriptsize
  \begin{tabular}{M{0.5cm}M{0.8cm}M{0.2cm}L{3.0cm}M{0.8cm}} 
\toprule
 \multicolumn{4}{c}{Aggregation}  &  Acc \\
\midrule
A: & avg. & + & class-balanced thresh. &   50.3  \\
B: & avg. & + & thresh. &  55.9   \\
C: & thresh. & + & avg. &  \textbf{60.8} \\

\bottomrule
\end{tabular}
}
{%
  \caption{Temporal aggregation of frame-level pseudo labels into video-level pseudo labels on EADs $\rightarrow$ HMDB51 split 1.}%
  \label{tab:frame_label_aggregation}
}

\ffigbox{%
  \includegraphics[width=0.45\textwidth]{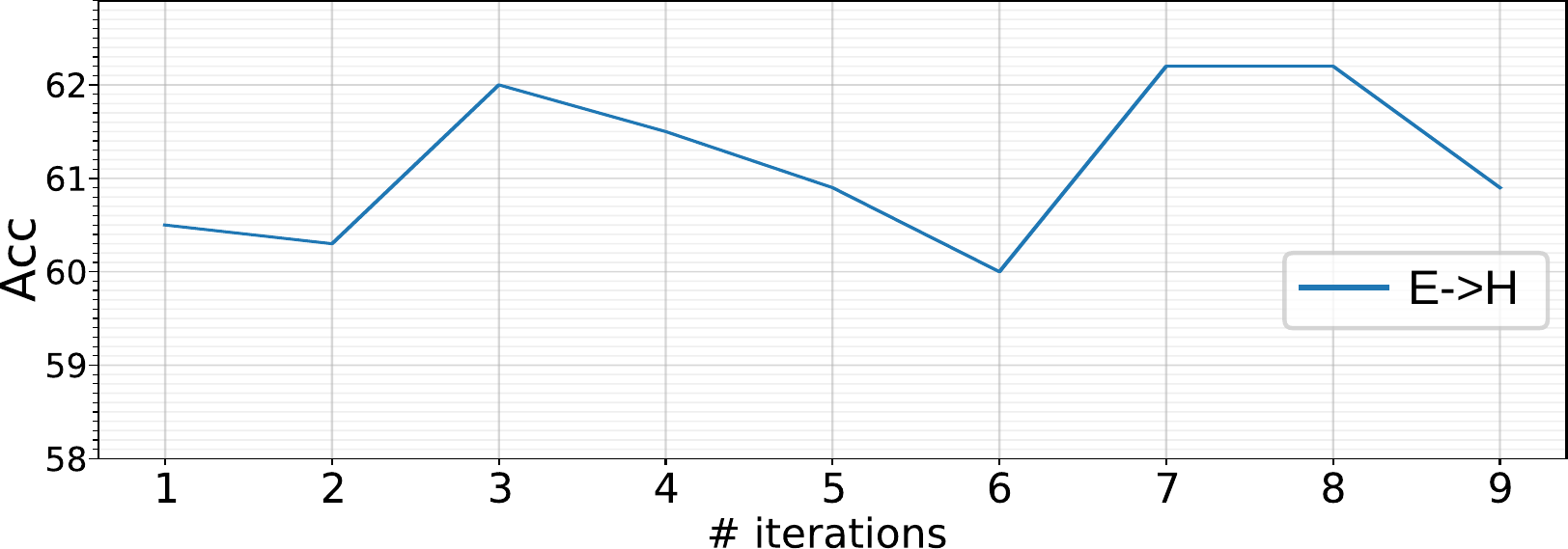}
}{%

  \caption{Performance of multiple iterations of CycDA (stage 3 and 4) on EADs $\rightarrow$ HMDB51. Average on 3 splits.}%
  \label{fig:n_iters_vs_performance_e2h}
}
\end{floatrow}
\end{figure}

\textbf{Temporal aggregation of frame-level pseudo labels}. Before video model training, we temporally aggregate frame-level pseudo labels. Here we compare three strategies of temporal aggregation in Table~\ref{tab:frame_label_aggregation}. Frame-level thresholding followed by temporal averaging (case C) outperforms the other two cases which perform temporal averaging before video-level thresholding. Frame-level thresholding filters out frames of low confidence scores and effectively increases the accuracy of video pseudo labels.


\textbf{Number of iterations}. Finally, we illustrate the performance after several iterations on EADs$\rightarrow$HMDB51 in Fig.~\ref{fig:n_iters_vs_performance_e2h}. It shows that within the first five iterations, performing CycDA iteratively results in a slight increase of performance. Further, within all the 9 iterations, CycDA delivers relatively stable performance, with the result fluctuating between 60.0\% and 62.2\% without dropping below the ablated alternatives shown in Table~\ref{tab:stage_wise_ablation}.

\section{Conclusions}
We presented CycDA to address the image-to-video adaptation problem. We proposed to alternately perform spatial domain alignment to address the domain shift between images and target frames, and spatio-temporal learning to bridge the modality gap. Our evaluations across several benchmarks and datasets demonstrate that CycDA exploits the large informativity in the source images for enhanced performance on the target video classifier.

\myparagraph{Acknowledgements} This work was partially funded by the Austrian Research Promotion Agency (FFG) project 874065 and by the Christian Doppler Laboratory for Semantic 3D Computer Vision, funded in part by Qualcomm Inc.

\appendix
\section*{Supplementary}

For additional insights into our unsupervised cycle domain adaptation approach (CycDA), we first analyze the frame thresholding in Sec.~\ref{sec:frame_thresholding} and provide additional baselines of image-to-video, frame-to-video and frame\&video-to-video adaptation in Sec.~\ref{sec:additonal_baseline}. We illustrate the performance of multiple iterations of CycDA on two adaptation datasets in Sec.~\ref{sec:multiple_iterations}.
Then, we demonstrate that the domain alignment improves the learned feature representations by searching nearest neighbors to target query frames in the source domain in Sec.~\ref{sec:nearest_neighbor}.
In Sec.~\ref{sec:cf_mat}, we inspect the confusion matrices of our generated pseudo labels.
Finally, we analyze potential failure cases of CycDA in Sec.~\ref{sec:failure_cases}.

\section{Frame Thresholding}\label{sec:frame_thresholding}
In the spatio-temporal learning stage (Sec.~3.2 in the main paper), we train a video model with pseudo labeled data in the target domain. To remove pseudo labels with low confidence, we set the confidence threshold $\delta_p$ such that $p\times$100\% of videos with highest confidence remain after the thresholding. 

We determine the maximum of the frame-level confidence scores within each video, and sort the maximum frame-level confidence scores of all videos in a descending order. The ordered score sequence is denoted as $[s_1, s_2, ..., s_{N}  ]$. For $\delta_p\in[s_{ \lfloor p\cdot N\rfloor  }, s_{ \lfloor p\cdot N \rfloor +1 }]$, there are $ \lfloor p\cdot N \rfloor$ videos left after thresholding. We set $\delta_p = (s_{ \lfloor p\cdot N\rfloor  }  + s_{ \lfloor p\cdot N \rfloor +1 }) /2 $.

\begin{figure}[!h] 
\centering
\includegraphics[width=0.5\textwidth]{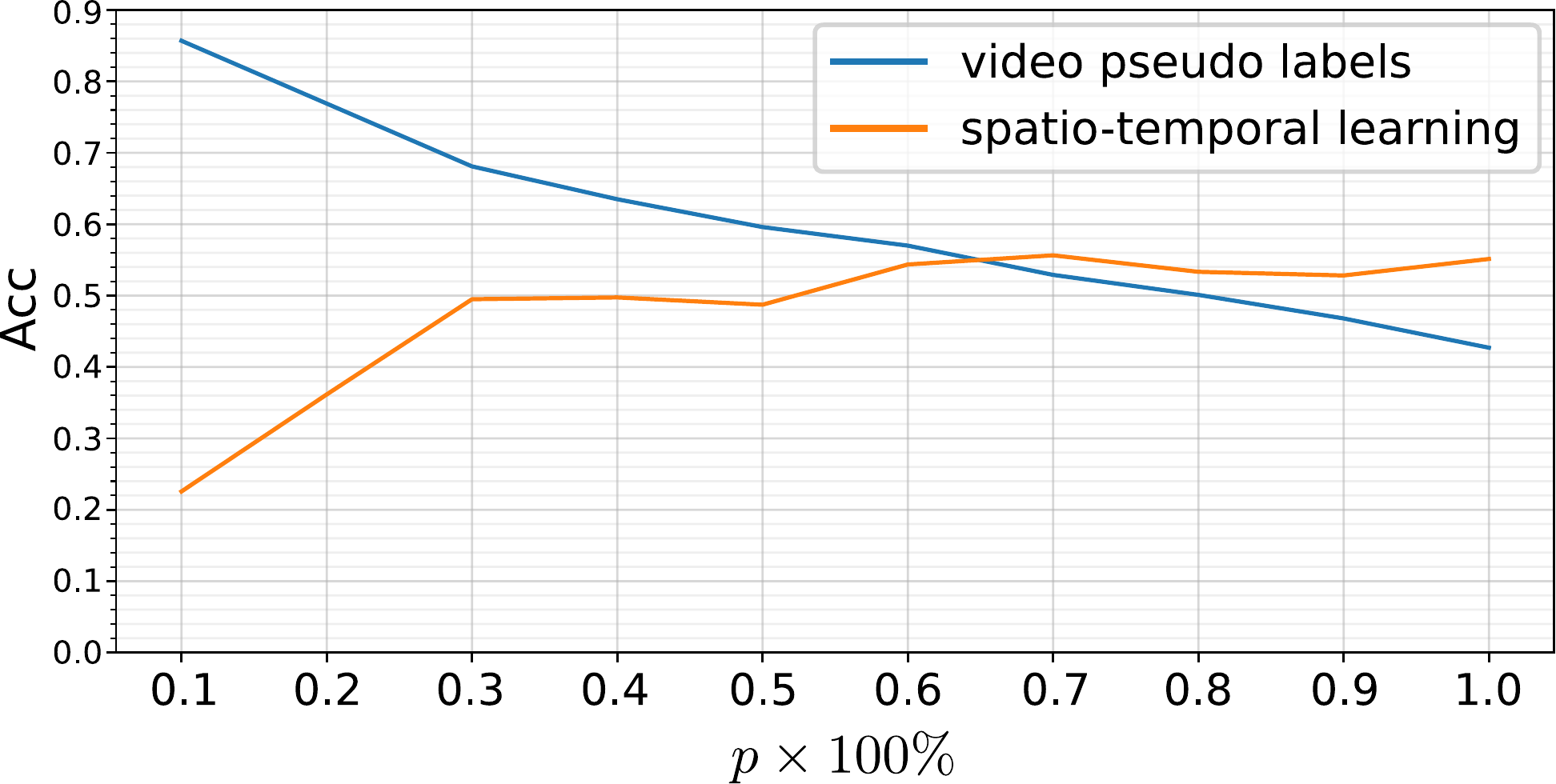}
\caption{Accuracy of video pseudo labels and the spatio-temporal learning (Stage 2) performance w.r.t. different thresholding percentages $p\times$100\% on EADs $\rightarrow$ HMDB51.  
}
\label{fig:thresh_p_vs_performance}
\end{figure}

We illustrate the accuracy of video pseudo labels and the spatio-temporal learning (Stage 2) performance in Fig.~\ref{fig:thresh_p_vs_performance}. Intuitively, with $p$ increasing, the accuracy of video pseudo labels drops, from 85.7\% ($p=10\%$) to 42.7\% ($p=100\%$). For $p$ smaller than 60\%, the spatio-temporal learning performance suffers due to the insufficient amount of training data, in spite of the higher accuracy of pseudo labels. For $p$ between 70\% and 100\%, the performance remains fairly stable at an accuracy of approximately 55\%. We set $p$ to 70\% for Stage~2 in all the experiments in the main paper. As Stage~3 further improves the pseudo labels transferred to Stage~4, we increase $p$ by 10\% to 80\% for the spatio-temporal learning in Stage~4 heuristically.

\section{Additional baselines}\label{sec:additonal_baseline}
\subsection{Image-to-video adaptation from BU101 to UCF-HMDB}\label{sec:from_bu_to_ucfhmdb}
In Sec.~4.4 of the main paper, we sample 50 web images per class from the 12 classes in BU101 and adapt to the UCF-HMDB videos. The BU101 dataset has around 230 images per class on average.
Here, we evaluate the image-to-video adaptation with a varying amount of web images.
For this, we sample different numbers of web images per class from BU101 and also compare our CycDA with video-to-video adaptation approaches (on UCF-HMDB) in Table~\ref{tab:from_bu101_to_ucfhmdb}. Apparently, increasing the number of web images on BU101 leads to a significant performance boost.The image-to-video adaptation with 230 images per class (82.2\% for BU$\rightarrow$H and 97.9\% for BU$\rightarrow$U) outperforms most state-of-the-art video-to-video adaptation approaches. The performance of BU$\rightarrow$U even exceeds the supervised target model. This demonstrates that CycDA can exploit the large informativity in the web images for enhanced performance on the target video classifier. 

\subsection{Frame-to-Video and Frame\&Video-to-Video Adaptation}
We add the baselines of \textit{frame-to-video} in the case of 1 frame and 5 frames from each source video in Table~\ref{tab:from_bu101_to_ucfhmdb}. Furthermore, we add a new setting of \textit{frame\&video-to-video} adaptation: using 5 frames from each source video as the source image data (instead of web images), together with the videos as the source video data, to adapt to target videos with domain shift. Both, using more frames and using frame together with videos, lead to further improvements. 
\begin{table}[!tb]
\scriptsize
\centering
\begin{tabular}{M{0.2cm}M{2.3cm}M{1.8cm}M{3cm}M{2cm}M{1cm}M{1cm}} 
\toprule
\multicolumn{2}{c}{\multirow{2}*{ DA setting}} & \multirow{2}*{method}  & \multicolumn{2}{c}{source data} & \multirow{2}*{U$\rightarrow$H} & \multirow{2}*{H$\rightarrow$U}  \\
\cmidrule{4-5}
~ & ~ & ~  & BU web image (\# images per class) & videos (U or H) in \% & ~ & ~ \\
\midrule
\multirow{11}*{A:} & \multirow{11}*{video-to-video} & AdaBN~\cite{li2018adaptive} &  - & 100\% & 75.5  &  77.4 \\
~ & ~ & MCD~\cite{saito2018maximum} &  - & 100\% & 74.4  &  79.3 \\
~ &  ~ & TA\textsuperscript{3}N~\cite{chen2019temporal}  & - & 100\% & 78.3  &  81.8 \\
~ & ~ & ABG~\cite{luo2020adversarial} &   - & 100\% & 79.1  &  85.1 \\
~ & ~ & TCoN~\cite{pan2020adversarial} &   - & 100\% & \textbf{87.2}  &  89.1 \\
~ & ~ & DANN~\cite{ganin2016domain} &   - & 100\%  & 80.7 & 88.0 \\
~ & ~ & TA\textsuperscript{3}N~\cite{chen2019temporal}   & - & 100\%  & 81.4 & 90.5 \\
~ & ~ & SAVA~\cite{choi2020shuffle}   & - & 100\%  & 82.2 & 91.2 \\
~ & ~ & MM-SADA~\cite{munro2020multi}  & - & 100\% & 84.2  &  91.1 \\
~ & ~ & CrossModal~\cite{kim2021learning}  & - & 100\% & 84.7  &  92.8 \\
~ & ~ & CoMix~\cite{sahoo2021contrast}  & - & 100\%  &86.7 & \textbf{93.9}   \\
\midrule

\multirow{3}*{B:} & \multirow{3}*{\shortstack{image-to-video}} & \multirow{3}*{CycDA} &  50 & - & 77.8 & 88.6 \\
~ & ~ & ~ & 100 & - & 81.7 & 93.9 \\
~ & ~ & ~  &  230 & - & \textbf{82.2} & \textbf{97.9} \\
\midrule
\multirow{2}*{C:} & \multirow{2}*{\shortstack{frame-to-video}} & \multirow{2}*{CycDA} &  - & 1 frame & 83.3 & 80.4 \\
~ & ~ & ~ & - & 5 frames & 84.4 & 84.6 \\
\midrule
D: & frame\&video-to-video  & CycDA & - & 5~frames+videos & 85.6 & 87.9 \\
\midrule
\multicolumn{3}{c}{supervised target} &  - & - & 94.4 & 97.0 \\
\bottomrule
\end{tabular}
\vspace{2mm}
\caption{Results of image-to-video adaptation (from images in BU101 to videos on UCF-HMDB) in comparison to video-to-video adaptation approaches. }
\label{tab:from_bu101_to_ucfhmdb}
\end{table}

\section{Multiple Iterations}\label{sec:multiple_iterations}
\begin{figure}[!h] 
\centering
\includegraphics[width=0.5\textwidth]{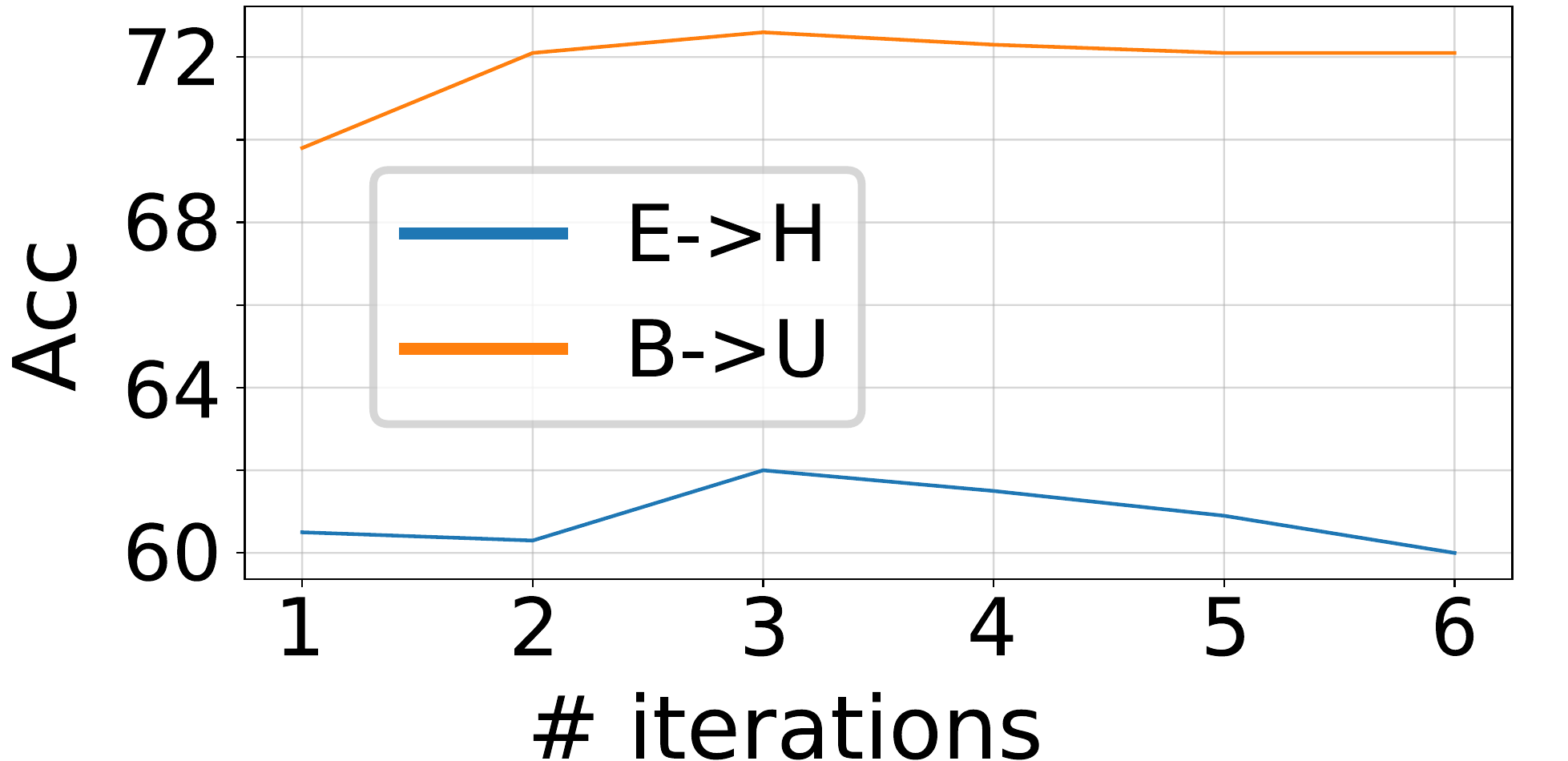}
\caption{Multiple iterations of CycDA on E$\rightarrow$H and B$\rightarrow$U.
}
\label{fig:iterations_vs_performance}
\end{figure}

We observed that the fluctuation effect of performance among multiple iterations is more pronounced for small datasets with large domain shift (such as E$\rightarrow$H), which are more prone to overfitting on either the image or the video side. To verify this, we compare the behaviour on E$\rightarrow$H with B$\rightarrow$U for 6 iterations in Fig.~\ref{fig:iterations_vs_performance}. While in both cases the peak is reached after three iterations, the performance on the large-scale B$\rightarrow$U remains more stable.

\section{Nearest Neighbor Search in Image Feature Space}\label{sec:nearest_neighbor}
In Sec.~4.5 in the main paper, we perform a stage-wise ablation study to show how each stage contributes to the performance boost. In order to further demonstrate how the domain alignment is improved, we use the sampled target frames as query and search for their nearest neighbor (NN) source web images.
For this, we use the following three image feature representations: (i)~source-only, (ii)~class-agnostic domain alignment (CycDA stage~1) and (iii)~class-aware domain alignment (CycDA stage 3).
The t-SNE visualizations of these 3 image feature spaces are shown in the main paper in Figure 3(a)--(c).
For further investigation, we sample query frames from target videos and show their 5 nearest neighbor source web images in Fig.~\ref{fig:nn_fig1} and~\ref{fig:nn_fig2}. In each subfigure, the 3 rows show the nearest neighbor results in the image feature space of the source-only model (1st row), after CycDA stage 1 class-agnostic alignment (2nd row), and after CycDA stage~3 class-aware alignment (3rd row).

We see that the nearest neighbors of the target query frame in the source-only feature space are from different categories, due to the large domain shift between source and target distributions before alignment. After class-agnostic domain alignment, the amount of source nearest neighbors from the same category slightly increases. After class-aware domain alignment in stage~3, most of the nearest neighbors are from the same category. In Fig.~\ref{fig:nn_fig2}(b), where the target query frame shows a child \textit{clapping}, both source-only and class-agnostic alignment result in several source nearest neighbors which show children, but none of these belong to the \textit{clap} category. On the contrary, class-aware alignment leads to nearest neighbors with non-baby content in the correct category. 
This indicates our effective category-level alignment which semantically gathers target frames with source data of the same category, instead of simply aligning images in terms of styles and appearance. Similar examples can be found in Fig.~\ref{fig:nn_fig1}(c) and~(d).

\begin{figure}[!h] 
\centering
\includegraphics[width=\textwidth]{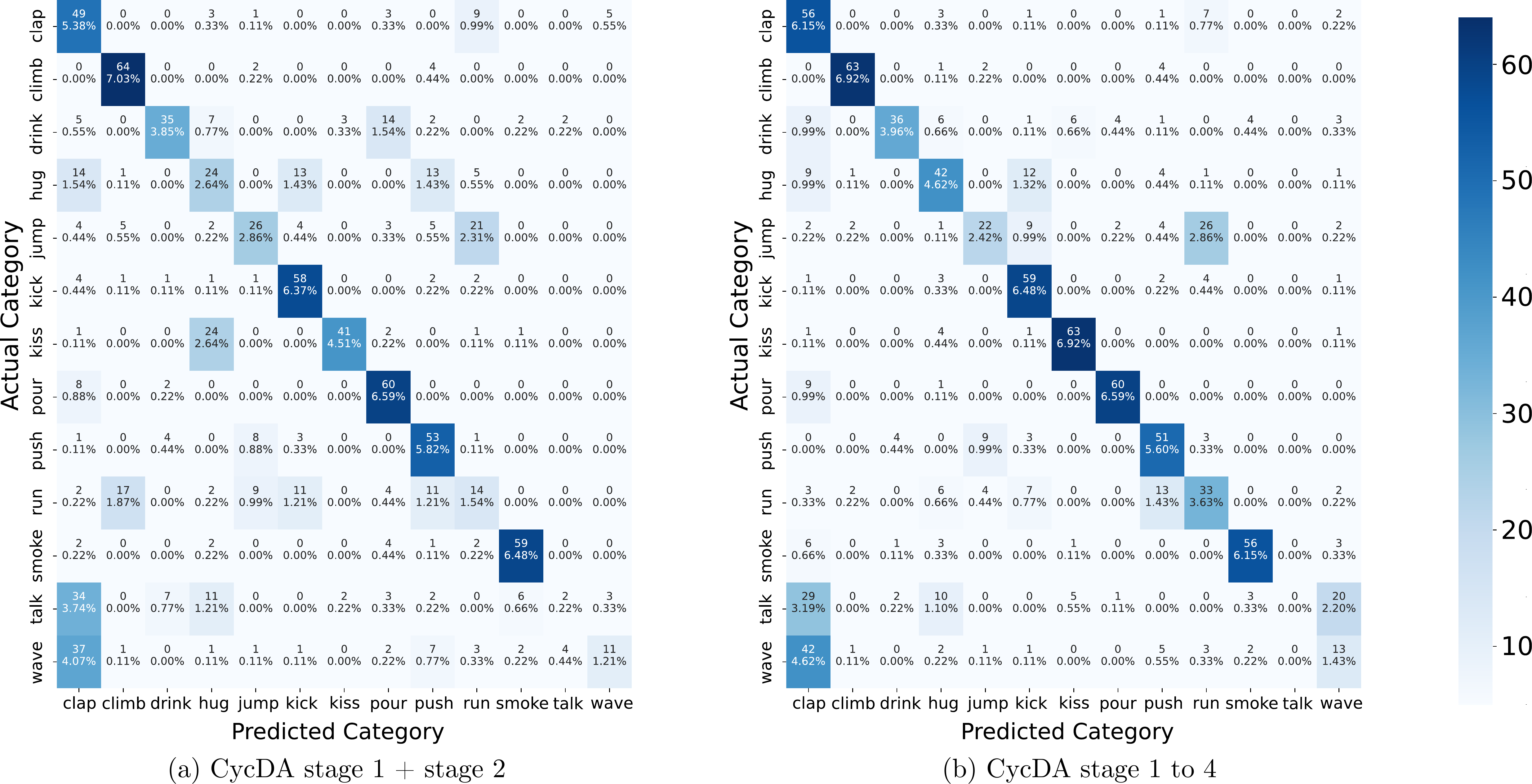}
\caption{Confusion matrices of video pseudo labels on target videos after (a) CycDA stage 2 and (b) CycDA stage 4 for 12 classes on EADs $\rightarrow$ HMDB51.
Best viewed on screen.
}
\label{fig:cf_mat}
\end{figure}




\begin{figure}[!h] 
\centering
\includegraphics[width=\textwidth]{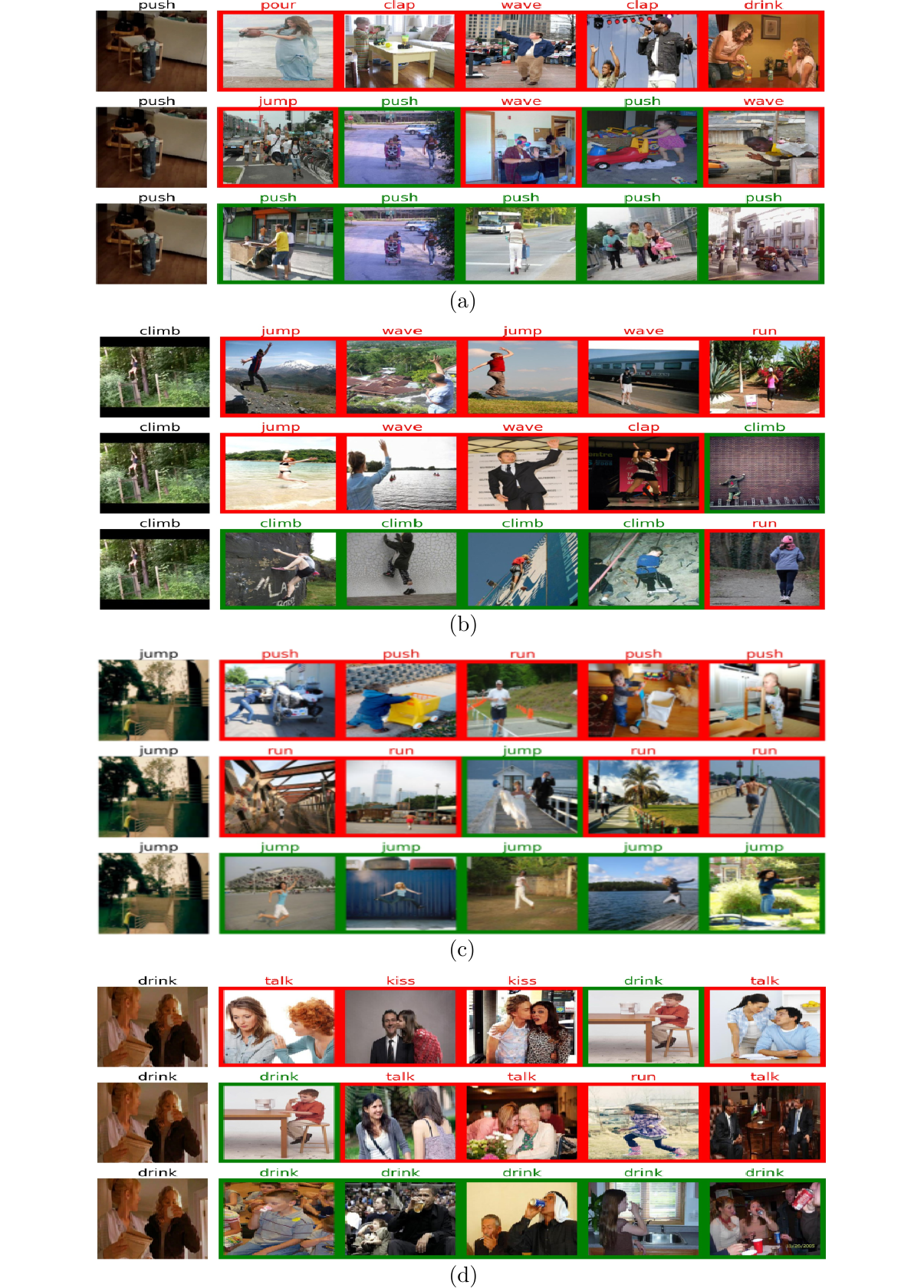}
\caption{Target video frame and its 5 nearest neighbor (NN) web images in the source domain. Each 3-row group subfigure displays the NN search in image feature space of source only (1st row), CyDA stage 1 class-agnostic alignment (2nd row), CycDA stage 3 class-aware alignment (3rd row). The image border color indicates NN in same (green) or different (red) category. }
\label{fig:nn_fig1}
\end{figure}

\begin{figure}[!h] 
\centering
\includegraphics[width=\textwidth]{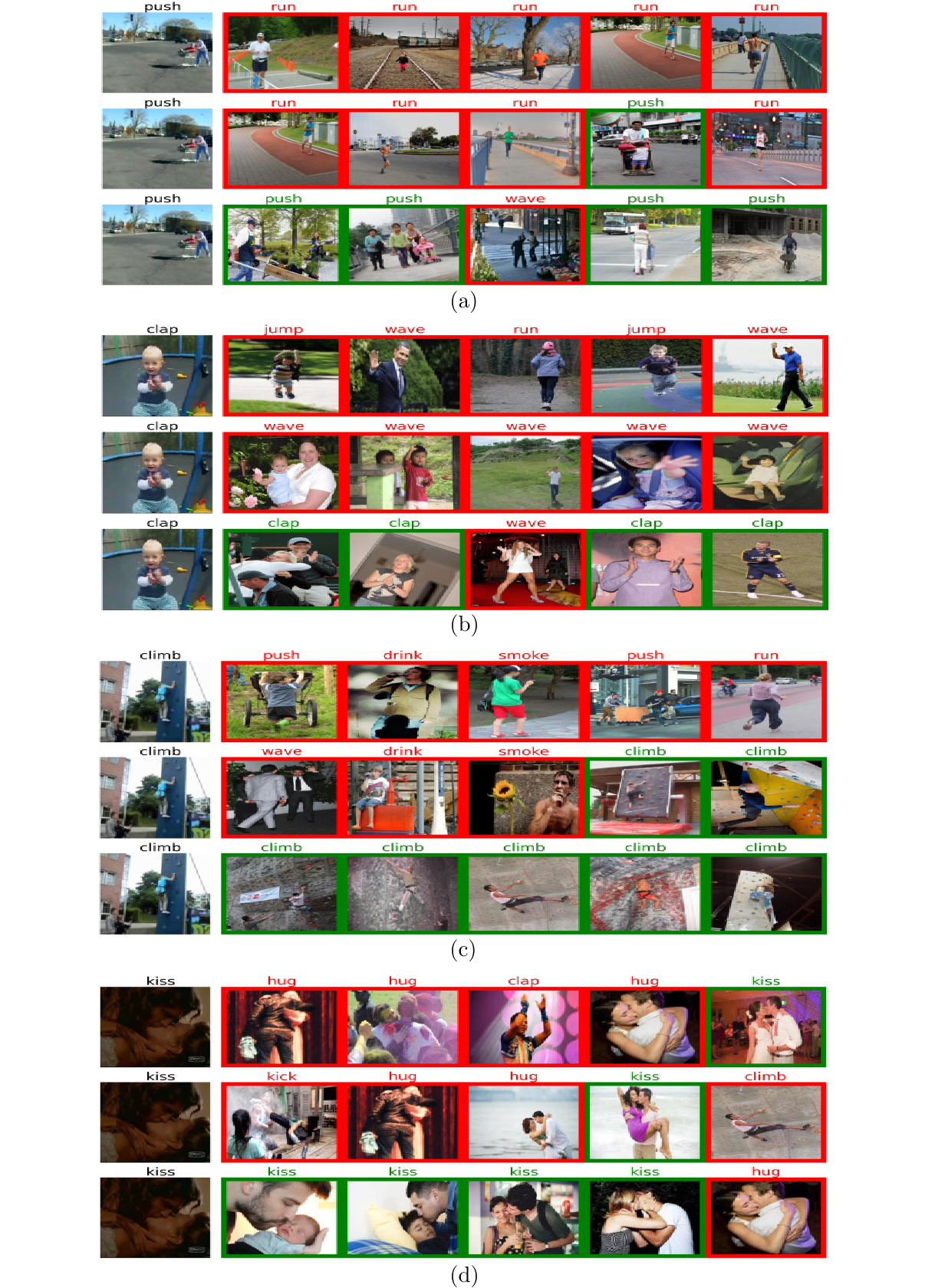}
\caption{Target video frame and its 5 nearest neighbor (NN) web images in the source domain. Each 3-row group subfigure displays the NN search in image feature space of source only (1st row), CyDA stage 1 class-agnostic alignment (2nd row), CycDA stage 3 class-aware alignment (3rd row). The image border color indicates NN in same (green) or different (red) category. }
\label{fig:nn_fig2}
\end{figure}


\section{Confusion Matrix}\label{sec:cf_mat}
Complementing our pseudo label analysis (Fig.~4(a) in the main paper), we further illustrate the confusion matrices of the video pseudo labels on the target videos after CycDA stage 2 (Fig.~\ref{fig:cf_mat}(a)) and CycDA stage 4 (Fig.~\ref{fig:cf_mat}(b)).
Clearly, Fig.~\ref{fig:cf_mat}(b) improves along the diagonal on the difficult classes (\eg \textit{run, hug, kiss}). 
Furthermore, Fig.~\ref{fig:cf_mat}(b) shows less confusion in comparison to Fig.~\ref{fig:cf_mat}(a), \eg between \textit{kiss} and \textit{hug}, \textit{drink} and \textit{pour}, or \textit{run} and \textit{climb}. 
Our complete CycDA (with stages~3 and~4) contributes to a significant performance boost with less confusion among categories.

The remaining confusion in Fig.~\ref{fig:cf_mat}(b) is due to low inter-class variation (\cf Sec.~\ref{sec:failure_cases}), \eg between \textit{wave} and \textit{clap}, or \textit{jump} and \textit{run}. The most difficult category is \textit{talk} (with zero accuracy), which is defined on EADs as the interaction between two or more subjects. This interaction is captured by our model and provides a strong bias for classification. On HMDB51, however, most \textit{talk} videos contain only a single subject speaking and thus, capture no interaction.

\section{Failure Cases}\label{sec:failure_cases}
\begin{figure}[!h] 
\centering
\includegraphics[width=\textwidth]{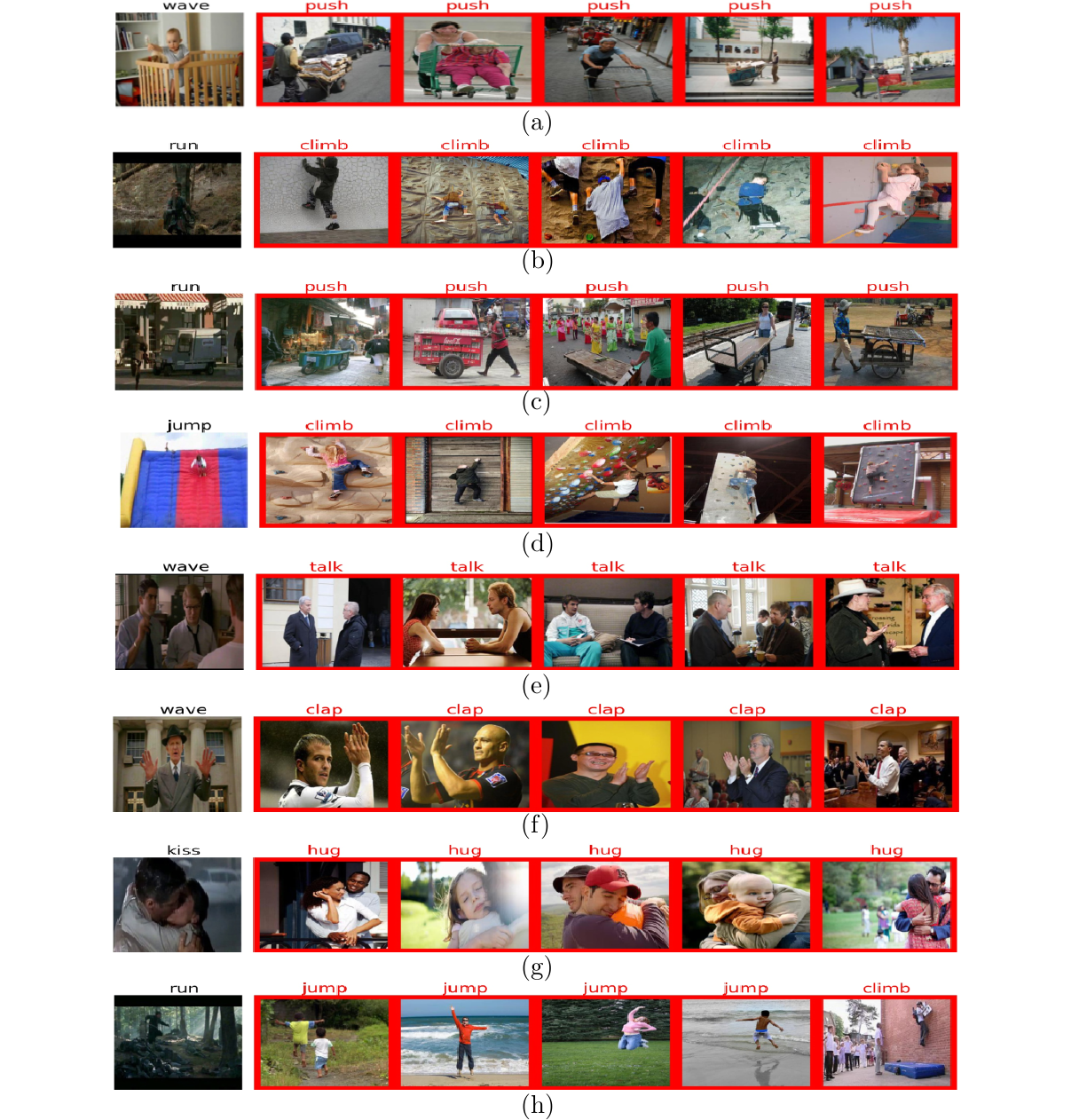}
\caption{Failure cases in search of 5 nearest neighbors (NN) in the image feature space of CycDA stage 3 class-aware domain alignment. The image border color indicates NN in same (green) or different (red) category.}
\label{fig:failure_cases}
\end{figure}

To analyze the limitations of CycDA, Fig.~\ref{fig:failure_cases} illustrates failure cases of the nearest neighbor search in the image feature space after the step~3 class-aware domain alignment.
%
The majority of our failures can be attributed to unusual backgrounds (where the background has a high visual similarity to the typical scenario of another action category) and ambiguous actions (low inter-class variation). 
For example, Fig.~\ref{fig:failure_cases}(a) shows a baby \textit{waving} in a crib, where the crib has a similar pattern to carts that are typically found in the \textit{push} category. Similarly, \textit{running} in front of rocks (Fig.~\ref{fig:failure_cases}(b)) or \textit{running} towards a vehicle (Fig.~\ref{fig:failure_cases}(c)) is grouped to \textit{climb} or \textit{push} respectively. \textit{Jumping} on the slope of a bouncy castle (Fig.~\ref{fig:failure_cases}(d)) looks visually similar to the scene of \textit{climb}.

Fig.~\ref{fig:failure_cases}(e)--(h) demonstrate category confusion due to low inter-class variation. For example, \textit{waving} while \textit{talking} to another subject occurs frequently in the web images annotated as \textit{talk} (Fig.~\ref{fig:failure_cases}(e)). \textit{Waving} with both hands is visually similar to \textit{clap} (Fig.~\ref{fig:failure_cases}(f)). \textit{Kissing} while hugging can be confused with \textit{hugging} only (Fig.~\ref{fig:failure_cases}(g)). \textit{Running} can look like \textit{jumping} when the subject is in the air (Fig.~\ref{fig:failure_cases}(h)).
These ambiguous actions can also be seen from the confusion matrices in Fig.~\ref{fig:cf_mat}.
Although spatio-temporal learning on a video model improves the inference of motion-based actions on the target domain, these visually similar or ambiguous actions still pose a challenge.
Deriving motion information from source web images might be a solution to this issue. 

\bibliographystyle{splncs04}
\bibliography{egbib}

\begin{thebibliography}{10}
\providecommand{\url}[1]{\texttt{#1}}
\providecommand{\urlprefix}{URL }
\providecommand{\doi}[1]{https://doi.org/#1}

\bibitem{carreira2017quo}
Carreira, J., Zisserman, A.: Quo vadis, action recognition? a new model and the
  kinetics dataset. In: CVPR. pp. 6299--6308 (2017)

\bibitem{chen2021spatial}
Chen, J., Wu, X., Hu, Y., Luo, J.: Spatial-temporal causal inference for
  partial image-to-video adaptation. In: AAAI. vol.~35, pp. 1027--1035 (2021)

\bibitem{chen2019temporal}
Chen, M.H., Kira, Z., AlRegib, G., Yoo, J., Chen, R., Zheng, J.: Temporal
  attentive alignment for large-scale video domain adaptation. In: ICCV. pp.
  6321--6330 (2019)

\bibitem{chen2020action}
Chen, M.H., Li, B., Bao, Y., AlRegib, G., Kira, Z.: Action segmentation with
  joint self-supervised temporal domain adaptation. In: CVPR. pp. 9454--9463
  (2020)

\bibitem{chen2020simple}
Chen, T., Kornblith, S., Norouzi, M., Hinton, G.: A simple framework for
  contrastive learning of visual representations. In: ICML. pp. 1597--1607.
  PMLR (2020)

\bibitem{choi2020unsupervised}
Choi, J., Sharma, G., Chandraker, M., Huang, J.B.: Unsupervised and
  semi-supervised domain adaptation for action recognition from drones. In:
  WACV. pp. 1717--1726 (2020)

\bibitem{choi2020shuffle}
Choi, J., Sharma, G., Schulter, S., Huang, J.B.: Shuffle and attend: Video
  domain adaptation. In: ECCV. pp. 678--695. Springer (2020)

\bibitem{deng2009imagenet}
Deng, J., Dong, W., Socher, R., Li, L.J., Li, K., Fei-Fei, L.: Imagenet: A
  large-scale hierarchical image database. In: CVPR. pp. 248--255. Ieee (2009)

\bibitem{duan2020omni}
Duan, H., Zhao, Y., Xiong, Y., Liu, W., Lin, D.: Omni-sourced webly-supervised
  learning for video recognition. In: ECCV. pp. 670--688. Springer (2020)

\bibitem{feichtenhofer2020x3d}
Feichtenhofer, C.: X3d: Expanding architectures for efficient video
  recognition. In: CVPR (2020)

\bibitem{gan2016webly}
Gan, C., Sun, C., Duan, L., Gong, B.: Webly-supervised video recognition by
  mutually voting for relevant web images and web video frames. In: ECCV. pp.
  849--866. Springer (2016)

\bibitem{gan2017deck}
Gan, C., Sun, C., Nevatia, R.: Deck: Discovering event composition knowledge
  from web images for zero-shot event detection and recounting in videos. In:
  AAAI. vol.~31 (2017)

\bibitem{gan2016you}
Gan, C., Yao, T., Yang, K., Yang, Y., Mei, T.: You lead, we exceed: Labor-free
  video concept learning by jointly exploiting web videos and images. In: CVPR.
  pp. 923--932 (2016)

\bibitem{ganin2015unsupervised}
Ganin, Y., Lempitsky, V.: Unsupervised domain adaptation by backpropagation.
  In: ICML. pp. 1180--1189. PMLR (2015)

\bibitem{ganin2016domain}
Ganin, Y., Ustinova, E., Ajakan, H., Germain, P., Larochelle, H., Laviolette,
  F., Marchand, M., Lempitsky, V.: Domain-adversarial training of neural
  networks. JMLR  \textbf{17}(1),  2096--2030 (2016)

\bibitem{guo2018curriculumnet}
Guo, S., Huang, W., Zhang, H., Zhuang, C., Dong, D., Scott, M.R., Huang, D.:
  Curriculumnet: Weakly supervised learning from large-scale web images. In:
  ECCV. pp. 135--150 (2018)

\bibitem{he2016deep}
He, K., Zhang, X., Ren, S., Sun, J.: Deep residual learning for image
  recognition. In: CVPR. pp. 770--778 (2016)

\bibitem{jamal2018deep}
Jamal, A., Namboodiri, V.P., Deodhare, D., Venkatesh, K.: Deep domain
  adaptation in action space. In: BMVC. vol.~2, p.~5 (2018)

\bibitem{kae2020image}
Kae, A., Song, Y.: Image to video domain adaptation using web supervision. In:
  WACV. pp. 567--575 (2020)

\bibitem{karpathy2014large}
Karpathy, A., Toderici, G., Shetty, S., Leung, T., Sukthankar, R., Fei-Fei, L.:
  Large-scale video classification with convolutional neural networks. In:
  CVPR. pp. 1725--1732 (2014)

\bibitem{kay2017kinetics}
Kay, W., Carreira, J., Simonyan, K., Zhang, B., Hillier, C., Vijayanarasimhan,
  S., Viola, F., Green, T., Back, T., Natsev, P., et~al.: The kinetics human
  action video dataset. arXiv preprint arXiv:1705.06950  (2017)

\bibitem{kim2021learning}
Kim, D., Tsai, Y.H., Zhuang, B., Yu, X., Sclaroff, S., Saenko, K., Chandraker,
  M.: Learning cross-modal contrastive features for video domain adaptation.
  In: ICCV. pp. 13618--13627 (2021)

\bibitem{kuehne2011hmdb}
Kuehne, H., Jhuang, H., Garrote, E., Poggio, T., Serre, T.: Hmdb: a large video
  database for human motion recognition. In: ICCV. pp. 2556--2563. IEEE (2011)

\bibitem{li2017attention}
Li, J., Wong, Y., Zhao, Q., Kankanhalli, M.S.: Attention transfer from web
  images for video recognition. In: ACM Multimedia. pp.~1--9 (2017)

\bibitem{li2018adaptive}
Li, Y., Wang, N., Shi, J., Hou, X., Liu, J.: Adaptive batch normalization for
  practical domain adaptation. Pattern Recognition  \textbf{80},  109--117
  (2018)

\bibitem{liu2021cycle}
Liu, H., Wang, J., Long, M.: Cycle self-training for domain adaptation. arXiv
  preprint arXiv:2103.03571  (2021)

\bibitem{liu2020deep}
Liu, Y., Lu, Z., Li, J., Yang, T., Yao, C.: Deep image-to-video adaptation and
  fusion networks for action recognition. TIP  \textbf{29},  3168--3182 (2019)

\bibitem{luo2020adversarial}
Luo, Y., Huang, Z., Wang, Z., Zhang, Z., Baktashmotlagh, M.: Adversarial
  bipartite graph learning for video domain adaptation. In: ACM Multimedia. pp.
  19--27 (2020)

\bibitem{ma2017less}
Ma, S., Bargal, S.A., Zhang, J., Sigal, L., Sclaroff, S.: Do less and achieve
  more: Training cnns for action recognition utilizing action images from the
  web. Pattern Recognition  \textbf{68},  334--345 (2017)

\bibitem{van2008visualizing}
Van~der Maaten, L., Hinton, G.: Visualizing data using t-sne. JMLR
  \textbf{9}(11) (2008)

\bibitem{munro2020multi}
Munro, J., Damen, D.: Multi-modal domain adaptation for fine-grained action
  recognition. In: CVPR. pp. 122--132 (2020)

\bibitem{pan2020adversarial}
Pan, B., Cao, Z., Adeli, E., Niebles, J.C.: Adversarial cross-domain action
  recognition with co-attention. In: AAAI. vol.~34, pp. 11815--11822 (2020)

\bibitem{sahoo2021contrast}
Sahoo, A., Shah, R., Panda, R., Saenko, K., Das, A.: Contrast and mix: Temporal
  contrastive video domain adaptation with background mixing. In: NeurIPS
  (2021)

\bibitem{saito2018maximum}
Saito, K., Watanabe, K., Ushiku, Y., Harada, T.: Maximum classifier discrepancy
  for unsupervised domain adaptation. In: CVPR. pp. 3723--3732 (2018)

\bibitem{soomro2012ucf101}
Soomro, K., Zamir, A.R., Shah, M.: Ucf101: A dataset of 101 human actions
  classes from videos in the wild. arXiv preprint arXiv:1212.0402  (2012)

\bibitem{sun2015temporal}
Sun, C., Shetty, S., Sukthankar, R., Nevatia, R.: Temporal localization of
  fine-grained actions in videos by domain transfer from web images. In: ACM
  Multimedia. pp. 371--380 (2015)

\bibitem{tanisik2016facial}
Tanisik, G., Zalluhoglu, C., Ikizler-Cinbis, N.: Facial descriptors for human
  interaction recognition in still images. Pattern Recognition Letters
  \textbf{73},  44--51 (2016)

\bibitem{wang2017untrimmednets}
Wang, L., Xiong, Y., Lin, D., Van~Gool, L.: Untrimmednets for weakly supervised
  action recognition and detection. In: CVPR. pp. 4325--4334 (2017)

\bibitem{wang2021action}
Wang, Z., She, Q., Smolic, A.: Action-net: Multipath excitation for action
  recognition. In: CVPR (2021)

\bibitem{yang2020temporal}
Yang, C., Xu, Y., Shi, J., Dai, B., Zhou, B.: Temporal pyramid network for
  action recognition. In: CVPR (2020)

\bibitem{yang2018recognition}
Yang, J., Sun, X., Lai, Y.K., Zheng, L., Cheng, M.M.: Recognition from web
  data: A progressive filtering approach. TIP  \textbf{27}(11),  5303--5315
  (2018)

\bibitem{yao2011human}
Yao, B., Jiang, X., Khosla, A., Lin, A.L., Guibas, L., Fei-Fei, L.: Human
  action recognition by learning bases of action attributes and parts. In:
  ICCV. pp. 1331--1338. IEEE (2011)

\bibitem{yu2019exploiting}
Yu, F., Wu, X., Chen, J., Duan, L.: Exploiting images for video recognition:
  Heterogeneous feature augmentation via symmetric adversarial learning. TIP
  \textbf{28}(11),  5308--5321 (2019)

\bibitem{yu2018exploiting}
Yu, F., Wu, X., Sun, Y., Duan, L.: Exploiting images for video recognition with
  hierarchical generative adversarial networks. In: IJCAI (2018)

\bibitem{zhang2016semi}
Zhang, J., Han, Y., Tang, J., Hu, Q., Jiang, J.: Semi-supervised image-to-video
  adaptation for video action recognition. IEEE Trans Cybern  \textbf{47}(4),
  960--973 (2016)

\bibitem{zhang2020label}
Zhang, Y., Deng, B., Jia, K., Zhang, L.: Label propagation with augmented
  anchors: A simple semi-supervised learning baseline for unsupervised domain
  adaptation. In: ECCV. pp. 781--797. Springer (2020)

\bibitem{zhuang2017attend}
Zhuang, B., Liu, L., Li, Y., Shen, C., Reid, I.: Attend in groups: a
  weakly-supervised deep learning framework for learning from web data. In:
  CVPR. pp. 1878--1887 (2017)

\bibitem{zou2018unsupervised}
Zou, Y., Yu, Z., Kumar, B., Wang, J.: Unsupervised domain adaptation for
  semantic segmentation via class-balanced self-training. In: ECCV. pp.
  289--305 (2018)

\bibitem{zou2019confidence}
Zou, Y., Yu, Z., Liu, X., Kumar, B., Wang, J.: Confidence regularized
  self-training. In: ICCV. pp. 5982--5991 (2019)

\end{thebibliography}

\end{document}